\documentclass[10pt,journal,compsoc]{IEEEtran}
\usepackage{booktabs}
\usepackage{threeparttable}

\ifCLASSOPTIONcompsoc
  \usepackage[nocompress]{cite}
\else
  \usepackage{cite}
\fi
\ifCLASSINFOpdf
  \usepackage[pdftex]{graphicx}
   \graphicspath{{imgs/}}
\else
\fi
\usepackage{color}
\usepackage{amsmath}
\usepackage[cmintegrals]{newtxmath}
\usepackage[noend]{algpseudocode}
\usepackage{algorithm}
\begin{document}
%
\title{Monocular 3D Fingerprint\\Reconstruction and Unwarping}
%
%
%
%

\author{Zhe~Cui,
        Jianjiang~Feng,~\IEEEmembership{Member,~IEEE,}
        Jie~Zhou,~\IEEEmembership{Senior~Member,~IEEE}
\thanks{
Zhe Cui is with the School of Artificial Intelligence, Beijing University of Posts and Telecommunications, Beijing 100876, China, and Department of Automation, Beijing National Research Center for Information Science and Technology, Tsinghua University, Beijing 100084, China.

Jianjiang Feng and Jie Zhou are with Department of Automation, Beijing National Research Center for Information Science and Technology, Tsinghua University, Beijing 100084, China.

e-mail: cuizhe@bupt.edu.cn, jfeng@tsinghua.edu.cn, jzhou@tsinghua.edu.cn.

}
}

\IEEEtitleabstractindextext{%
\begin{abstract}
Compared with contact-based fingerprint acquisition techniques, contactless acquisition has the advantages of less skin distortion, larger fingerprint area, and hygienic acquisition. However, perspective distortion is a challenge in contactless fingerprint recognition, which changes ridge orientation, frequency, and minutiae location, and thus causes degraded recognition accuracy. We propose a learning based shape from texture algorithm to reconstruct a 3D finger shape from a single image and unwarp the raw image to suppress perspective distortion. Experimental results on contactless fingerprint databases show that the proposed method has high 3D reconstruction accuracy. Matching experiments on contactless-contact and contactless-contactless matching prove that the proposed method improves matching accuracy.
\end{abstract}

}

\maketitle

\IEEEdisplaynontitleabstractindextext

%
\IEEEpeerreviewmaketitle

\IEEEraisesectionheading{\section{Introduction}\label{sec:introduction}}
\IEEEPARstart{T}{raditional} fingerprint acquisition relies on pressing fingers against fingerprint acquisition devices to get ridge-valley images. But the collected fingerprints are often with distortion issues due to skin deformation, which degrades fingerprint recognition accuracy \cite{2009handbook}. In comparison, contactless fingerprint acquisition system provides a touch-free imaging approach, which does not have skin distortion, can obtain a fingerprint with larger area, and is more hygienic \cite{kumar2018contactless}.

Early researchers build camera systems to capture contactless fingerprints, which then go through conventional fingerprint recognition algorithms \cite{song2004new}\cite{lee2006study}\cite{hiew2007touch}\cite{kumar2011contactless}. However, as contactless fingerprints are directly captured by cameras, the illumination condition, camera resolution, focus clarity, and other imaging problems make contactless fingerprints of poorer quality than contact-based ones \cite{zhou2014performance}\cite{fiumara2018nist}. To solve quality issues, several deep learning approaches \cite{chopra2018unconstrained}\cite{lin2019cnn}\cite{tan2020towards} have been introduced recently and have gained promising results in dealing with various conditions of fingerprint quality. 

\begin{figure}[t]
\centering
\centerline{\includegraphics[width=\linewidth]{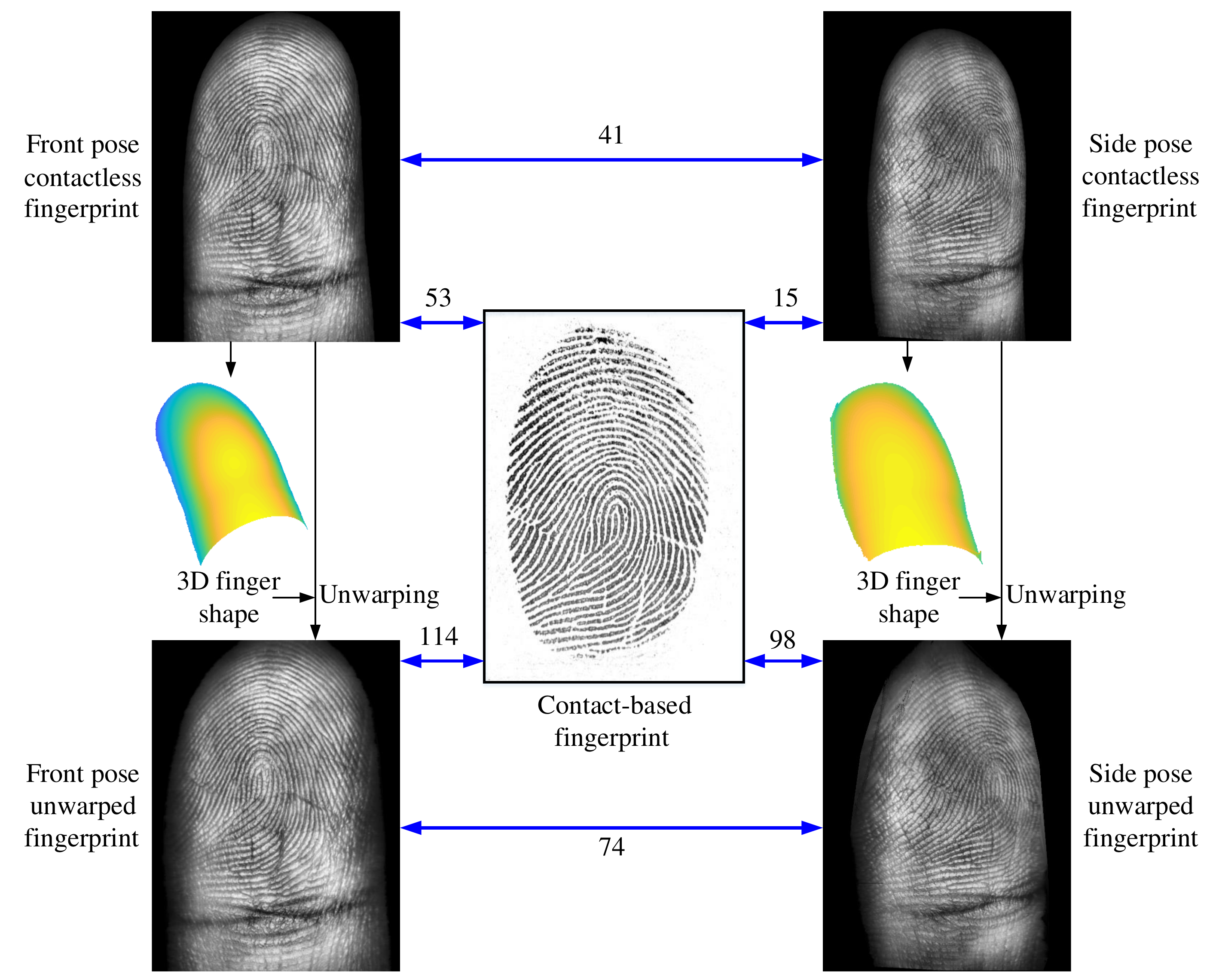}}
\caption{Given an original contactless fingerprint, the proposed algorithm reconstructs its 3D finger shape and then unwarps it. Using the unwarped images, matching scores of contactless-contact and contactless-contactless matching are improved significantly (53$\rightarrow$114, 15$\rightarrow$98, 41$\rightarrow$74). The numbers on lines are matching scores by a commercial fingerprint matcher, VeriFinger SDK 12.0 \cite{verifinger}.}
\label{fig:flowchart}
\end{figure}

Another major challenge of contactless fingerprints is that they usually have perspective distortion resulting from camera imaging. The ridge orientation, frequency, and minutiae location are changed while projecting a 3D finger onto a 2D plane, leading to challenges for traditional ridge enhancement, minutiae extraction, and fingerprint matching algorithms \cite{jain1997line}. As Fig. \ref{fig:flowchart} shows, perspective distortion degrades matching performances of both contactless-contact and contactless-contactless matching, and needs to be handled.

Some researchers consider the finger pose to solve perspective distortion issues \cite{labati2013contactless}\cite{tan2020towards}. The finger rotation angle is estimated from a contactless fingerprint, then a simulated finger model is rotated in 3D space and is reprojected to a 2D fingerprint to reduce view angle differences. Although such processing can improve matching accuracy, perspective distortion is not solved, as the rectified image is still a projection from the 3D model to the 2D plane. Dabouei {\emph{et al.}}\cite{Dabouei2019} consider perspective distortion as a problem similar to skin distortion \cite{si2015detection} and compute a thin-plate spline transformation model to undistort contactless fingerprints. But the unwarping process treats contactless fingerprints as 2D images, which lacks 3D geometric constraints.

Although there is great progress in contactless fingerprint matching, the problem of perspective distortion has not yet been solved, as there is currently no 3D fingerprint reconstruction algorithm that can directly estimate finger shape from a single contactless fingerprint image. Only by obtaining the 3D model of the finger surface can we explicitly model the perspective distortion and correct it.

The ridge-valley pattern of the human finger provides a natural texture clue for inferring 3D shape from the ridge pattern. But as fingerprint ridge frequency is not globally fixed and the ridges in contactless fingerprint images are unclear, especially in the finger border region, conventional shape-from-texture methods \cite{blostein1989shape} cannot achieve good 3D reconstruction results. 

This paper proposes a learning based 3D reconstruction and unwarping method for contactless fingerprints. To our knowledge, this is the first monocular 3D fingerprint reconstruction algorithm for contactless fingerprints that relies on only one fingerprint image. Our method infers 3D finger shape using fingerprint features learned from training samples. 

Fig. \ref{fig:flowchart} shows the impact of our algorithm on fingerprint matching performance. By reconstructing a 3D finger shape from a single contactless fingerprint and then unwarping the contactless fingerprint to suppress perspective distortion, the unwarped fingerprints have higher matching scores. Thus, the matching performance involving contactless fingerprints is improved.

We evaluate our method on three contactless fingerprint databases: PolyU Contactless 3D Fingerprint Database \cite{liu20143d}, UNSW 2D/3D Fingerprint Database \cite{zhou2014benchmark}, and PolyU Contactless 2D to Contact-based 2D Fingerprint Database \cite{lin2018matching} for evaluating 3D reconstruction performance and the contribution of our algorithm to matching performance. The reconstruction results show that the proposed reconstruction method reaches satisfactory reconstruction accuracy. The matching experiments show that the proposed unwarping method improves matching performance compared to raw images and gets state-of-the-art matching results.
\section{Related Work}\label{sec:related work}
Contactless fingerprint recognition methods can be categorized into 2D-based and 3D-based methods. The following is the review of related work of these two categories.
\subsection{Contactless 2D Fingerprint Recognition}\label{ssec:contactless_1}
To process contactless fingerprint images, early researchers directly use traditional fingerprint recognition methods \cite{song2004new}\cite{lee2006study}\cite{hiew2007touch}. Some develop specific devices to capture contactless fingerprints \cite{parziale2006surround}\cite{stein2012fingerphoto}\cite{sankaran2015smartphone}. 

But the accuracy of contactless fingerprint recognition is bothered by the image quality and perspective distortion. Some methods utilize high-resolution cameras for better image quality \cite{stein2012fingerphoto}\cite{sankaran2015smartphone}\cite{malhotra2017fingerphoto}, and a recent advance is introducing deep learning into contactless fingerprint processing \cite{labati2018novel}\cite{tan2020towards} and matching \cite{lin2017multi}\cite{lin2018matching}\cite{chopra2018unconstrained}\cite{lin2019cnn} to improve matching accuracy.

As for perspective distortion, pose constraint is considered for better results \cite{stein2012fingerphoto}\cite{labati2013contactless}\cite{tan2020towards}. A finger viewing angle is estimated to rotate the finger to the same frontal perspective angle in 3D space. Although normalizing finger pose is beneficial for fingerprint matching, the corrected fingerprint is still a 2D projection of a 3D model, and perspective distortion still exists.

Some methods \cite{Dabouei2019}\cite{grosz2022contact} treat perspective distortion as skin distortion and rectify distortion by estimating a distortion field \cite{si2015detection}. But this unwarping method is 2D-based and does not consider 3D finger shape explicitly.
\subsection{Contactless 3D Fingerprint Reconstruction and Recognition}\label{ssec:contactless_2}
A recent advance is to combine 2D contactless fingerprint images and 3D finger shapes for better recognition results. To obtain a 3D fingerprint, various 3D reconstruction methods have been proposed. In multi-view 3D reconstruction systems \cite{liu20143d}\cite{labati2015toward}, multiple cameras with known intrinsic and external parameters are required to obtain a 3D shape. 

In various other 3D sensing systems, such as shape from shading \cite{kumar2013towards}\cite{lin2017tetrahedron}, shape from focus \cite{abramovich2010mobile}, and shape from silhouette \cite{parziale2006surround}, multiple shots of fingerprint images are required to compute a 3D fingerprint and specially designed acquisition equipment is needed. Other 3D acquisition methods like structured light \cite{wang2010data}, ultrasound imaging \cite{baradarani2013resonance}, and lasers \cite{galbally2017full} are also applied to capture 3D fingerprints. 

Several researchers \cite{kumar2013towards}\cite{liu2015study}\cite{lin2017tetrahedron} further study the application of 3D fingerprint feature extraction and recognition after 3D reconstruction. The feature extraction and matching are conducted entirely in 3D space and don't consider the perspective distortion of fingerprint images.
\subsection{Summary}\label{ssec:contactless_3}
\newcommand{\tabincell}[2]{\begin{tabular}{@{}#1@{}}#2\end{tabular}}
\begin{table*}[htb]
\caption{A review of monocular contactless fingerprint researches.}
\centering
\begin{threeparttable}
\begin{tabular}{cclccc}
\toprule
Authors&Year&\tabincell{c}{Methods for Perspective Distortion}&Database&\tabincell{c}{CL-CL}&\tabincell{c}{CL-C}\\
\midrule

Lee \emph{et al.} \cite{lee2006study}&2006&Reject large view difference&Not public&--& \checkmark \\

Hiew \emph{et al.} \cite{hiew2007touch}&2007&\tabincell{c}{Only use center area}&Not public& \checkmark &--\\

Kumar \emph{et al.} \cite{kumar2011contactless}&2011&\tabincell{c}{Only use center area}&A& \checkmark &--\\

Stein \emph{et al.} \cite{stein2012fingerphoto}&2012&\tabincell{c}{Rotate fingerprints to same pose}&Not public& \checkmark &--\\

Labati \emph{et al.} \cite{labati2013contactless}&2013&\tabincell{c}{Rotate fingerprints to same pose}&Not public& \checkmark &--\\

Zhou \emph{et al.} \cite{zhou2014performance}&2014&--&B& \checkmark & \checkmark \\

Sankaran \emph{et al.} \cite{sankaran2015smartphone}&2015&--&C& \checkmark & \checkmark \\

Liu \emph{et al.} \cite{liu2016improved}&2016&--&Not public& \checkmark &--\\

Lin \emph{et al.} \cite{lin2018matching}&2018&\tabincell{c}{Training a correction model to rectify distortion}&\tabincell{c}{D}& -- & \checkmark \\

Lin \emph{et al.} \cite{lin2018contactless}&2018&--&\tabincell{c}{B, D}& -- &\checkmark\\

Dabouei \emph{et al.} \cite{Dabouei2019}&2019&\tabincell{c}{Training a transformation network to rectify distortion}&\tabincell{c}{B, E}& -- &\checkmark\\

Tan \emph{et al.} \cite{tan2020towards}&2020&\tabincell{c}{Rotate fingerprints to same pose}&\tabincell{c}{B, F}& \checkmark &--\\

S{\"o}llinger \emph{et al.} \cite{Sollinger2021}&2021&\tabincell{c}{Using a parametric model to unwarp}&Not public& -- &\checkmark\\

Grosz \emph{et al.} \cite{grosz2022contact}&2022&\tabincell{c}{Training a transformation network to rectify distortion}&\tabincell{c}{B, D, E}& -- &\checkmark\\

Proposed Method&2022&\tabincell{c}{3D to 2D unwarping}&\tabincell{c}{B, D, G}&\checkmark &\checkmark\\
\bottomrule
\end{tabular}
\begin{tablenotes}
\item Database A: PolyU Finger Image Database \cite{kumar2011contactless}
\item Database B: UNSW 2D/3D Fingerprint Database \cite{zhou2014benchmark}
\item Database C: IIITD Smartphone Database \cite{sankaran2015smartphone}
\item Database D: PolyU Contactless 2D to Contact-Based 2D Fingerprint Database \cite{lin2018matching}
\item Database E: ManTech Database \cite{ericson2015evaluation}
\item Database F: PolyU 3D Fingerprint Images Database \cite{lin2017tetrahedron}
\item Database G: PolyU Contactless 3D Fingerprint Database \cite{liu20143d}
\item CL-CL: Contactless-to-contactless matching
\item CL-C: Contactless-to-contact matching
\end{tablenotes}
\end{threeparttable}
\label{table:review}
\end{table*}
Table \ref{table:review} summarizes previous studies on monocular contactless fingerprint recognition, and our method falls into this category. As the table shows, existing contactless 2D fingerprint recognition methods either lack 3D information or do not address perspective distortion. Meanwhile, existing contactless 3D fingerprint reconstruction methods usually require bulky and expensive acquisition equipment, impeding the widespread use of the technology.

Therefore, a 3D fingerprint reconstruction and unwarping method is proposed in the paper for robust and low cost monocular contactless fingerprint processing.
\section{Fingerprint Reconstruction and Unwarping}\label{sec:method}
\begin{figure*}[!]
\centering
\centerline{\includegraphics[width=\linewidth]{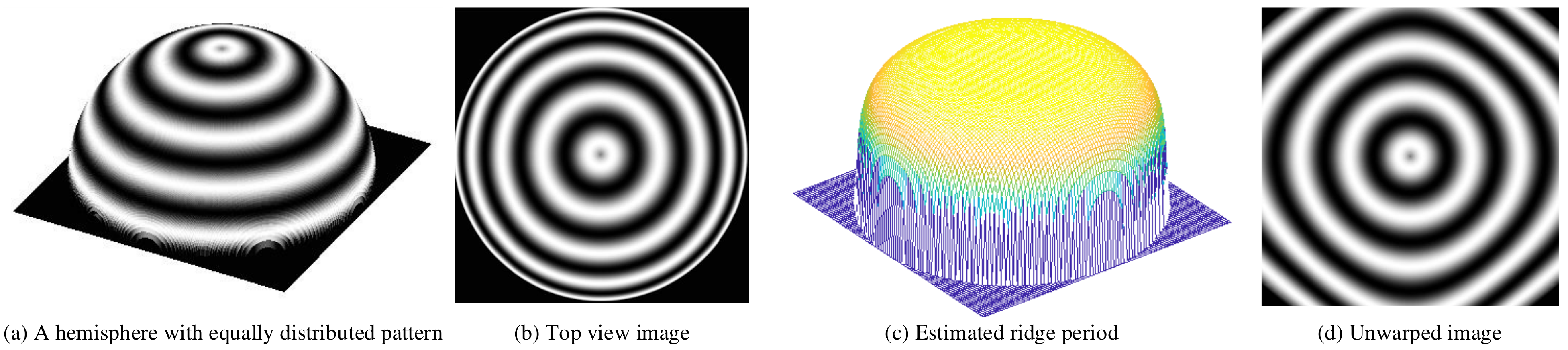}}
\caption{The basic methodology of our method. (a) is a hemisphere with equally distributed ridges along the radius. (b) is its projected image into a plane, and local ridge periods are changed due to perspective distortion. (c) is the estimated ridge period map from (b) using the short-time Fourier transform, where the change of period corresponds to a tilted surface. (d) is the unwarped image of (b) using an estimated 3D model, where the perspective distortion is reduced.}
\label{fig:idea}
\end{figure*}
\begin{figure*}[htb]
\centering
\centerline{\includegraphics[width=\linewidth]{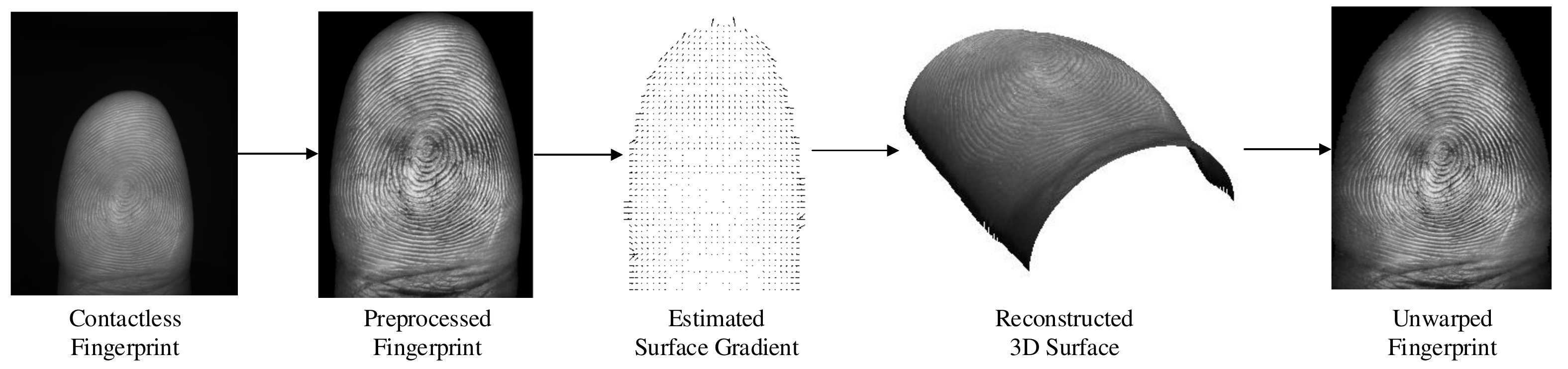}}
\caption{Flowchart of the proposed monocular 3D fingerprint reconstruction and unwarping algorithm.}
\label{fig:framework}
\end{figure*}
Fig. \ref{fig:idea} illustrates the basic idea of this paper using a simplified model. A hemisphere has an equally distributed ridge pattern along radius direction. But ridges in its 2D projection viewed from the top are not evenly distributed due to perspective distortion. The center area is less distorted because this area is perpendicular to the observer. The edge area has a reducing period and significant distortion because the tilt angle there is much larger. Therefore, the changes of the ridge period reflect the surface tilt angle towards the observer.

Similar to this simplified model, the local ridge period of contactless fingerprint clearly changes due to perspective distortion. The ridge period of the border region of the contactless fingerprint is clearly smaller than the center region due to the larger slant angle. The angle of the local surface can be inferred through the changes of the ridge period, which is a typical problem of shape from texture \cite{blostein1989shape}. 

Our method consists of 3D surface reconstruction from an input image and following image unwarping based on the reconstruction result. The whole algorithm mainly contains four steps: (1) contactless fingerprint preprocessing, (2) finger surface gradient estimation, (3) 3D surface reconstruction from the surface gradient, and (4) unwarping contactless fingerprint according to the reconstructed 3D shape. The flowchart of the proposed method is shown in Fig. \ref{fig:framework}.

\begin{figure*}[htb]
\centering
\centerline{\includegraphics[width=\linewidth]{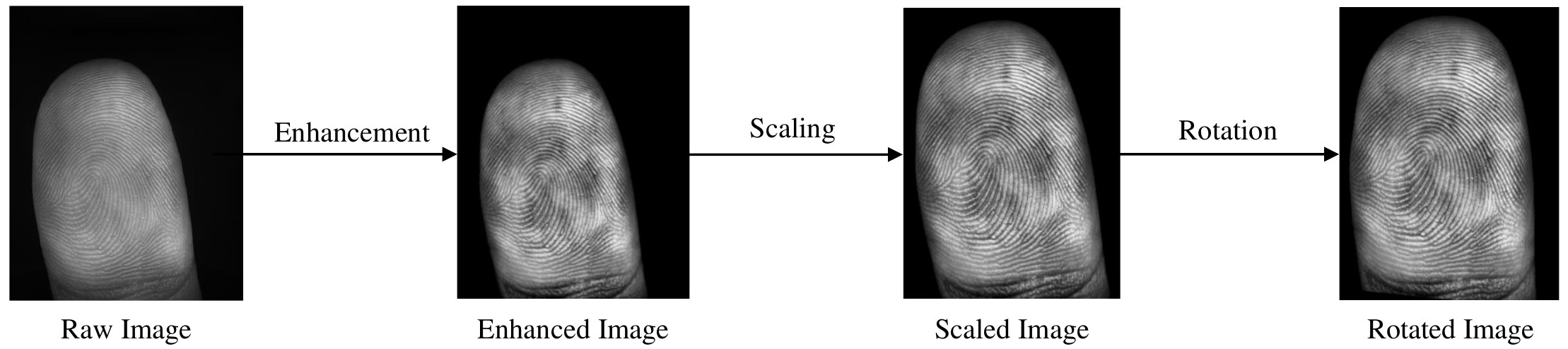}}
\caption{Procedure of image preprocessing, including enhancement, scaling, and rotation steps.}
\label{fig:preprocess}
\end{figure*}
\subsection{Image Preprocessing}\label{ssec:preprocess}
Raw contactless fingerprint images need to be preprocessed to reduce variations of input images. Following the processing method in \cite{Dabouei2019}, the preprocessing step alters images from three aspects: contrast, scale, and pose, which correspond to the three sub-modules illustrated in Fig. \ref{fig:preprocess} as enhancement, scaling, and rotation.

First, we enhance a raw fingerprint image to improve its ridge-valley contrast. Due to different lighting conditions, the image brightness may differ among different images. We first segment the finger regions by thresholding image intensities to find the contours of the finger. Then local adaptive histogram equalization is applied to enhance the raw image.

The next step is scaling. As monocular depth prediction has scale ambiguity issues \cite{eigen2014depth}, the scales of contactless images need to be settled. We use the average ridge period to determine the scale parameter. This is based on the observation that the central region of a contactless fingerprint barely has distortion, thus reflecting the actual scale of the original fingerprint. The mean ridge period of the human fingerprint is a constant number (about 0.5mm). Therefore, if we adjust the mean ridge period of the fingerprint image to a constant number, then the image-to-real ratio of fingerprint size is settled. After scaling, the ratio between the distance on the fingerprint image and the corresponding distance on the real finger is the same for all fingerprint images.

The detailed scaling steps are illustrated in Fig. \ref{fig:preprocess}. We find its geometric center as the ROI center point for each contactless fingerprint. Then a circle is drawn as the central region to cover about 20 percent of the total ROI region. The average ridge period is computed within the center area using a traditional algorithm \cite{hong1998fingerprint}. The fingerprint image is resized to change the mean ridge period to 10 pixels. By scaling images, the scale ambiguity issue is settled. We scale all ridge periods to 10 pixels because most flat fingerprints have a mean ridge period of 10 pixels under standard 500 ppi. Resizing to 10 pixels is beneficial for further matching contactless fingerprints with contact-based fingerprints.

\begin{figure}[!]
\centering
\centerline{\includegraphics[width=.5\linewidth]{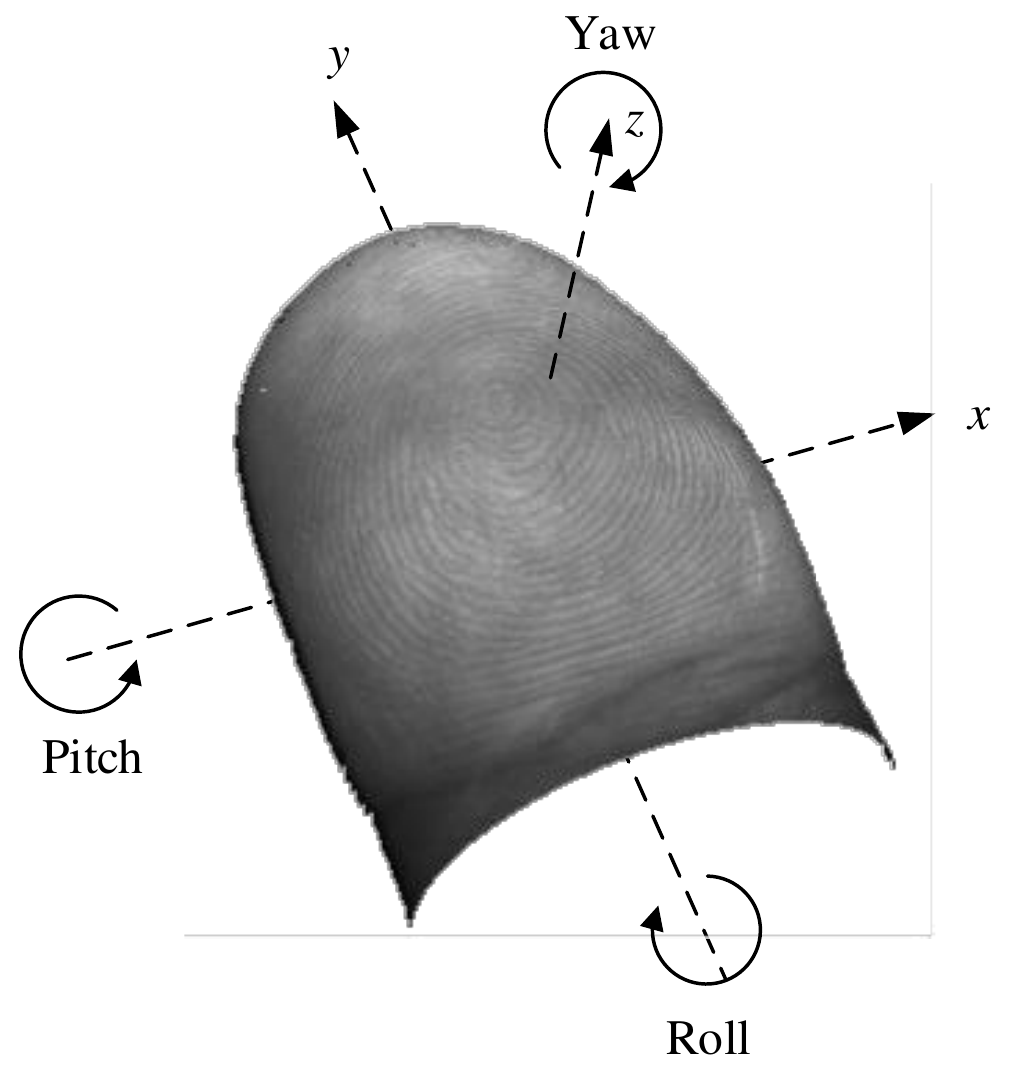}}
\caption{Three rotation angles of finger pose.}
\label{fig:angles}
\end{figure}

Finally, the pose variation is handled by rotation. As Fig. \ref{fig:angles} shows, the finger may rotate in three directions: \emph{Pitch} around $x$ axis, \emph{Roll} around $y$ axis, and \emph{Yaw} around $z$ axis. The pose correction step mainly deals with \emph{Yaw} angle, as \emph{Yaw} corresponds to a rotation in the 2D image plane and can be directly estimated from contactless images. \emph{Pitch} and \emph{Roll} angle variations are handled by data augmentation for training networks.

\emph{Yaw} angle is obtained by estimating the angle (or slope) of the centerline of the finger region. For each row in contactless fingerprint, we find its left and right contour, and locate its center point. We use the center points of all horizontal lines to compute a centerline by least-square fitting. Then, the angle between the centerline and vertical direction would be the \emph{Yaw} angle. A finger is rotated to make the centerline vertical.
\subsection{Surface Gradient Estimation}\label{ssec:netwrok}
This step estimates surface gradients for further 3D reconstruction using convolutional neural networks. We predict gradient instead of depth for three reasons:
\begin{itemize}
\item An evident pattern for contactless fingerprint is the change of ridge period from center to edge. This change directly corresponds to surface title degree or surface normal direction, which is equivalent to depth gradient.
\item Directly estimating depth has absolute value uncertainty. Namely, the absolute depth cannot be determined from the image. But gradient reflects the relative change. Thus, estimating the gradient can avoid this issue.
\item The following unwarping algorithm will need to calculate the geodesic distance on the 3D surface, which requires the gradient map.
\end{itemize}
\subsubsection{Network Structure}\label{sssec:structure}
\begin{figure*}[htb]
\centering
\centerline{\includegraphics[width=.7\linewidth]{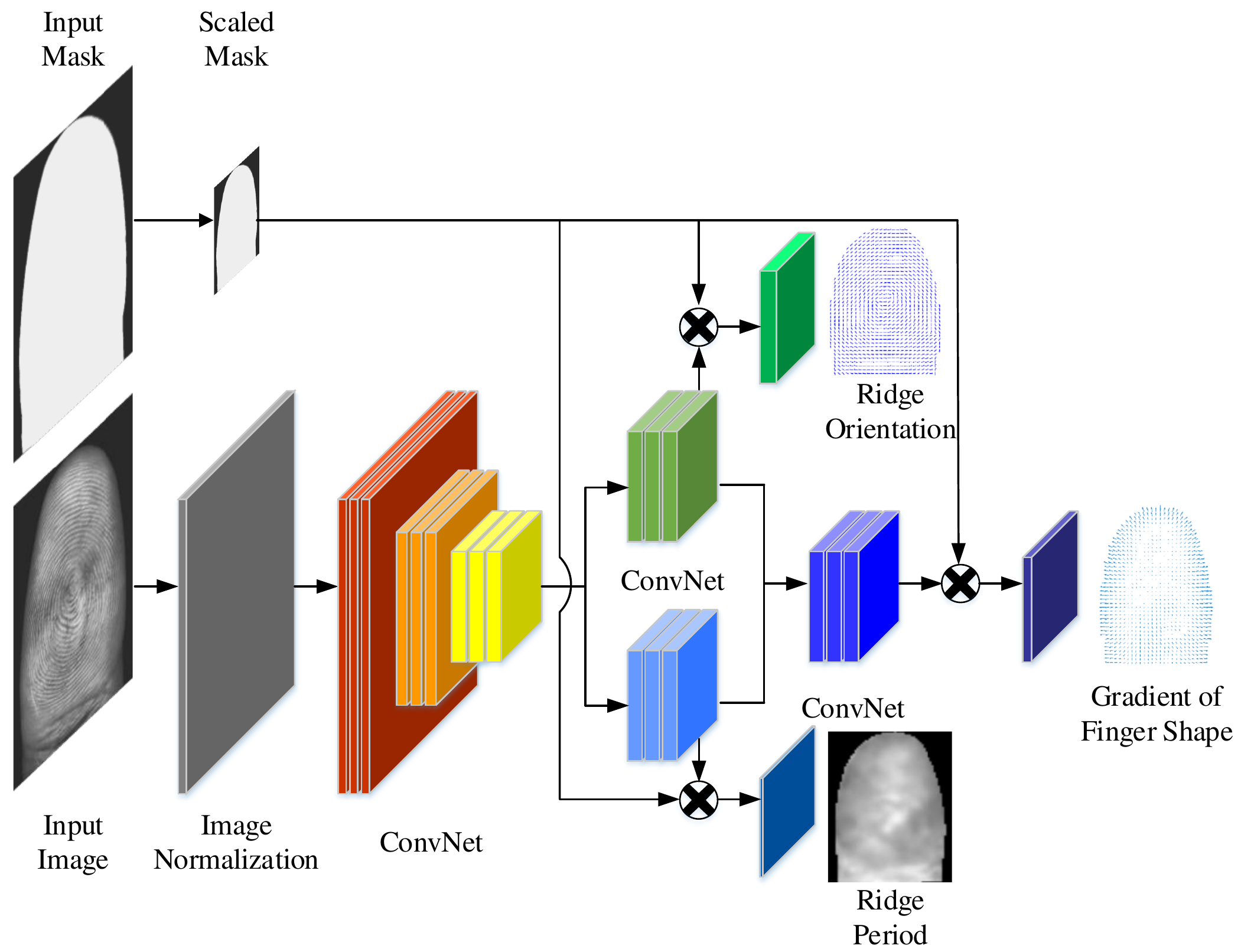}}
\caption{Structure of gradient estimation network.}
\label{fig:network}
\end{figure*}

Fig. \ref{fig:network} illustrates the structure of our gradient estimation network. The network takes the preprocessed image and segmentation mask as inputs, and outputs the orientation map, period map, and gradient map. The network is inspired by FingerNet \cite{fingernet2017}: the first part is image normalization to scale image intensities; the second part is orientation and period feature network, which consists of 3 convolution and max-pooling blocks to extract features of $1/8$ input size; the third part is to regress orientation and period maps separately; the final part is to regress gradients from orientation and period features. 

\begin{figure}[!]
\centering
\centerline{\includegraphics[width=.9\linewidth]{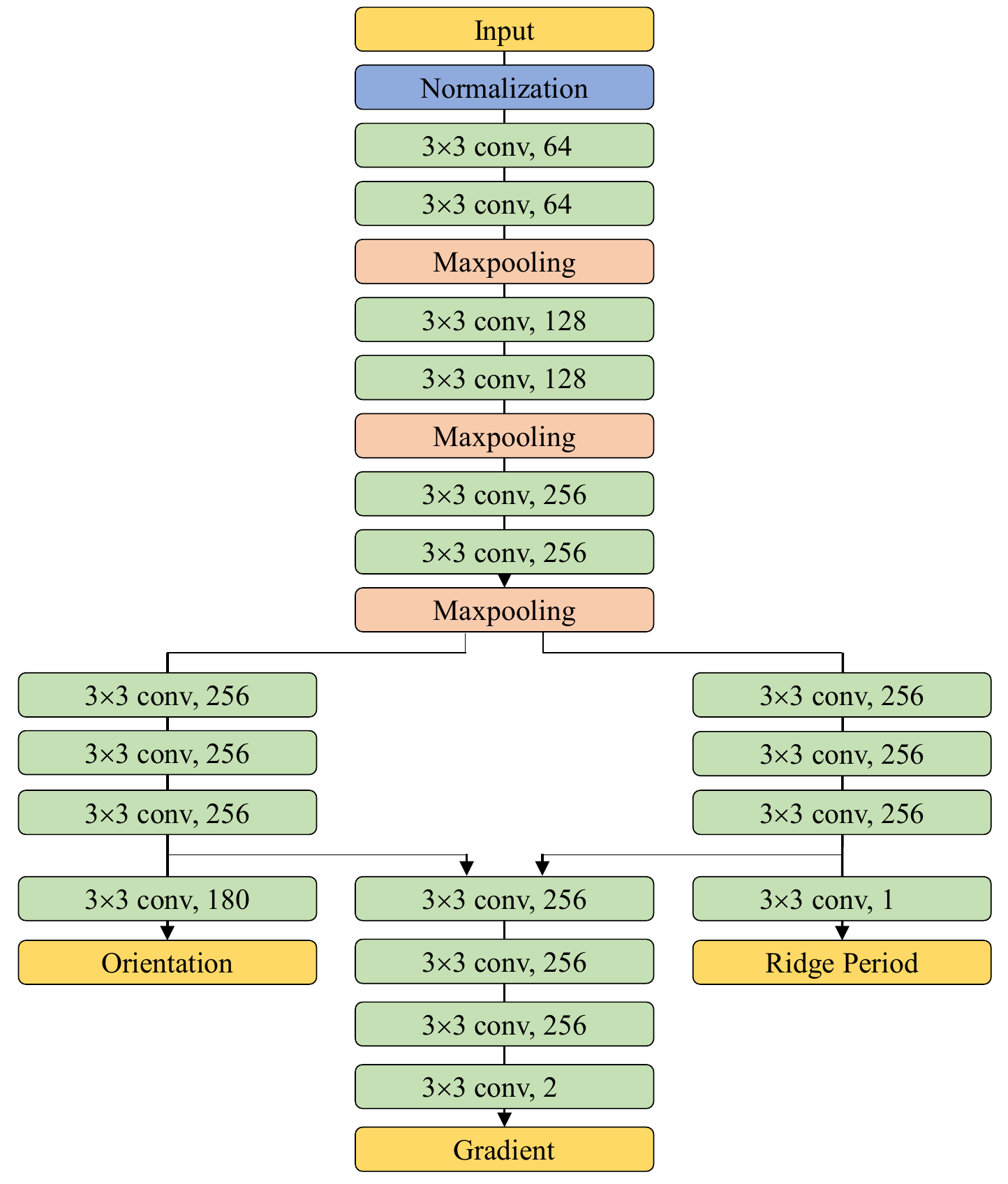}}
\caption{Details of network structure.}
\label{fig:network_structure}
\end{figure}
Fig. \ref{fig:network_structure} shows the details of the network structure. The first part, image normalization, is fulfilled at the end of image preprocessing, and is not included in the network structure. The second part is a VGG-like \cite{2015vgg} feature extraction block consisting of six convolution layers and three pooling layers. The extracted features are sent to orientation and period blocks simultaneously. Both blocks contain three convolution layers and a final regression layer. The features from orientation and period blocks are concatenated to the gradient regression block. The activation function is ReLU for each layer, except for the three output layers. Orientation output uses softmax activation, as it is a classification model with 180 categories. Period and gradient outputs use linear activation in regressing.

Noticing that extra mask input is added to activate output orientation, period, and gradient, since only the finger region is processed. The output orientation, period, and gradient maps have $1/8$ size of inputs during training, and the final outputs are interpolated to the original size when testing. 

The output orientation is an $N$-dimensional vector at each point $(x,y)$ standing for the probability distribution, where $N$ is the number of discrete angles of orientation. Here $N$ is set to $180$ in our experiment. The final orientation is a probability-weighted estimation \cite{kass1987analyzing}: 

\begin{equation}
\begin{aligned}
d_{\cos}(x,y)&=\frac{1}{N}\sum_{i=0}^{N-1}p(i,x,y)\cos(\frac{360}{N}i),\\
d_{\sin}(x,y)&=\frac{1}{N}\sum_{i=0}^{N-1}p(i,x,y)\sin(\frac{360}{N}i),\\
\text{Ori}(x,y)&=\frac{1}{2}\text{atan2}\left(d_{\sin}(x,y),d_{\cos}(x,y)\right).
\end{aligned}
\label{equ:ori}
\end{equation}

The output period is a scalar, while the gradient is 2-dimensional: $g_x(x,y)$ and $g_y(x,y)$ stand for gradients along $x$ and $y$ directions.
\subsubsection{Loss Definition}\label{sssec:losses}
There are three losses corresponding to three outputs: orientation loss $\mathcal{L}_{\text{Ori}}$, period loss $\mathcal{L}_{\text{Ped}}$, and gradient loss $\mathcal{L}_{\text{Grad}}$. 

Orientation loss $\mathcal{L}_{\text{Ori}}$ is composed of cross-entropy loss and coherence loss \cite{kass1987analyzing}:
\begin{equation}
\begin{aligned}
\mathcal{L}_{\text{Ori}}&=-\frac{1}{|M|}\sum_M\sum_{i=0}^{N-1}(o^{*}(i,x,y)\log(o(i,x,y))\\
&+(1-o^{*}(i,x,y))\log(1-o(i,x,y)))\\
&+\alpha\frac{|M|}{\sum_M\frac{\|(\overline{d}_{\cos}(x,y),\overline{d}_{\sin}(x,y))\|}{\overline{\text{Ori}}(x,y)}}-1.
\end{aligned}
\label{equ:ori_loss}
\end{equation}
$\overline{d}_{\cos}(x,y)$, $\overline{d}_{\sin}(x,y)$, and $\overline{\text{Ori}}(x,y)$ are the smoothing results of $d_{\cos}(x,y)$, $d_{\sin}(x,y)$, and $d(x,y)$ using a $3\times3$ all one kernel. $M$ is the input mask as activation, and $\sum_M{/}|M|$ means averaging within the mask region. $o(i,x,y)$ and $o^{*}(i,x,y)$ are estimation results and ground-truth respectively. $\alpha$ is the weight of coherence loss.

Period loss $\mathcal{L}_{\text{Ped}}$ is composed of regression loss and smoothing loss:
\begin{equation}
\begin{aligned}
\mathcal{L}_{\text{Ped}}&=-\frac{1}{|M|}\sum_M(p(x,y)-p^{*}(x,y))^2+\beta\|\nabla p(x,y)\|^2.
\end{aligned}
\label{equ:ped_loss}
\end{equation}
$p(x,y)$ and $p^{*}(x,y)$ are estimated period and ground-truth period. $\beta$ is the weight of smoothing loss. 

Gradient loss $\mathcal{L}_{\text{Grad}}$ is:
\begin{equation}
\begin{aligned}
\mathcal{L}_{\text{Grad}}&=-\frac{1}{\sum_M{w(x,y)}}\sum_Mw(x,y)\|g(x,y)-g^{*}(x,y)\|^2\\
&+\gamma\|\nabla g(x,y)\|^2.
\end{aligned}
\label{equ:grad_loss}
\end{equation}
$g(x,y)$ and $g^{*}(x,y)$ are estimated gradient and ground-truth gradient. $\gamma$ is the weight of smoothing loss. The gradient loss is similar to period loss, but with weight $w(x,y)$:
\begin{equation}
\begin{aligned}
w(x,y)=\exp\left(-\frac{\|g^{*}(x,y)\|}{\sigma}\right).
\end{aligned}
\label{equ:weight}
\end{equation}
$\sigma$ is the weight kernel. 

Because gradient data usually have larger value in the marginal region, and smaller value in the central region, giving all pixels the same weight using mean squared error in training leads to more attention on marginal area data than center area data. Therefore, we add weight here to balance the marginal area and central area. By setting larger values with smaller weights, their relative errors could be comparable.

The total loss is the sum of three losses:
\begin{equation}
\begin{aligned}
\mathcal{L}=\lambda_1\mathcal{L}_{\text{Ori}}+\lambda_2\mathcal{L}_{\text{Ped}}+\lambda_3\mathcal{L}_{\text{Grad}}.
\end{aligned}
\label{equ:loss}
\end{equation}
$\lambda_1$, $\lambda_2$, and $\lambda_3$ are the weights of three losses. 

All parameters used in the loss function are empirically determined. Table \ref{table:para} shows the parameter values during implementation.
\begin{table}[!]
\caption{Parameters used in loss function}
\centering
\begin{tabular}{cccccccc}
\toprule
Parameter&$\alpha$&$\beta$&$\gamma$&$\sigma$&$\lambda_1$&$\lambda_2$&$\lambda_3$\\
\midrule
Value&1&1&1&0.5&1&20&100\\
\bottomrule
\end{tabular}
\label{table:para}
\end{table}
\subsubsection{Real Training Data}\label{sssec:data}
\begin{table*}[!]
\caption{Contactless fingerprint databases used in this paper.}  
\centering
\begin{tabular}{cccc} 
\toprule
Database&\tabincell{c}{PolyU Contactless 3D\\Fingerprint Database \cite{liu20143d}} &\tabincell{c}{ UNSW 2D/3D\\Fingerprint Database \cite{zhou2014benchmark}}&\tabincell{c}{PolyU Contactless 2D to Contact-based\\2D Fingerprint Database \cite{lin2018matching}}\\
\midrule
Example & \tabincell{c}{\begin{minipage}{0.1\textwidth}\centerline{\includegraphics[scale=0.08]{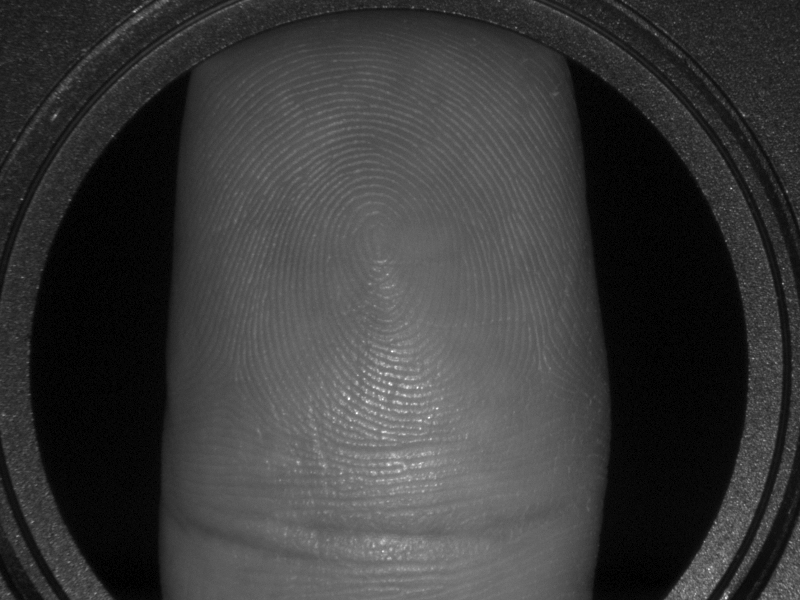}}\end{minipage} \\ \begin{minipage}{0.1\textwidth}\centerline{\includegraphics[scale=0.2]{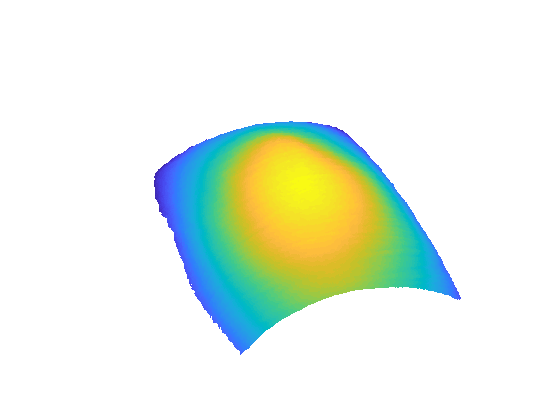}}\end{minipage}}&\tabincell{c}{ \begin{minipage}{0.1\textwidth}\centerline{\includegraphics[scale=0.05]{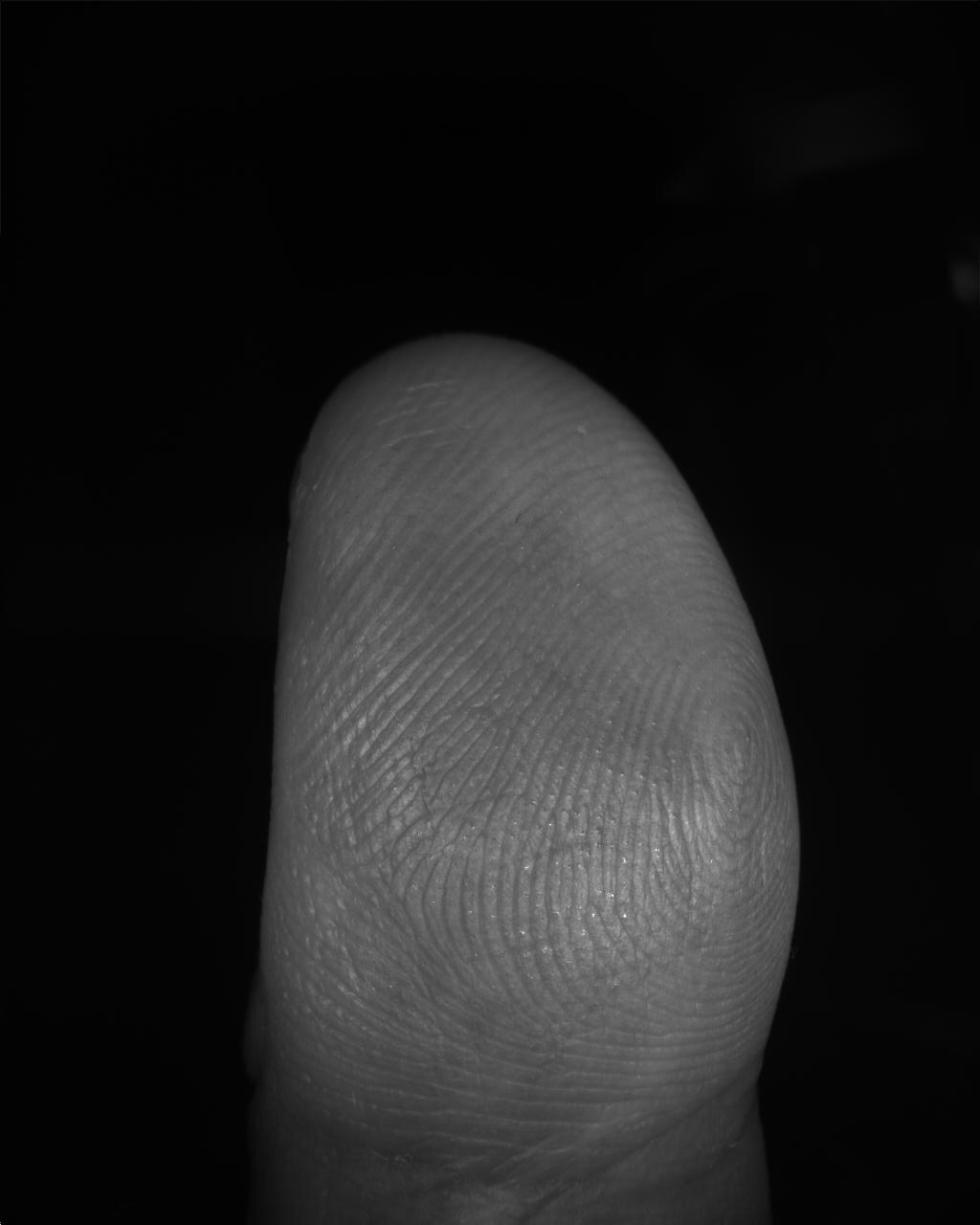}}\end{minipage} \begin{minipage}{0.1\textwidth}\centerline{\includegraphics[scale=0.05]{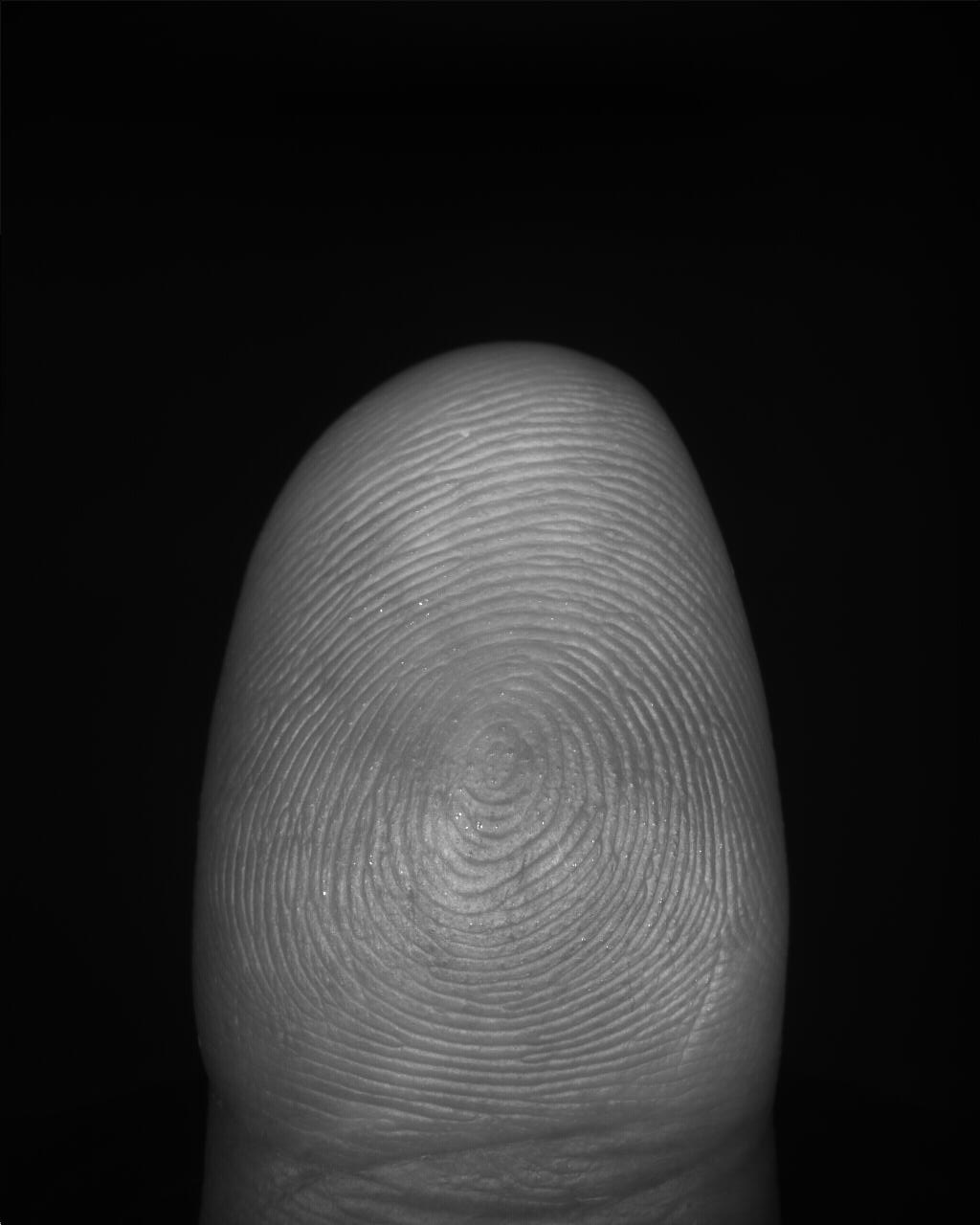}}\end{minipage} \begin{minipage}{0.1\textwidth}\centerline{\includegraphics[scale=0.05]{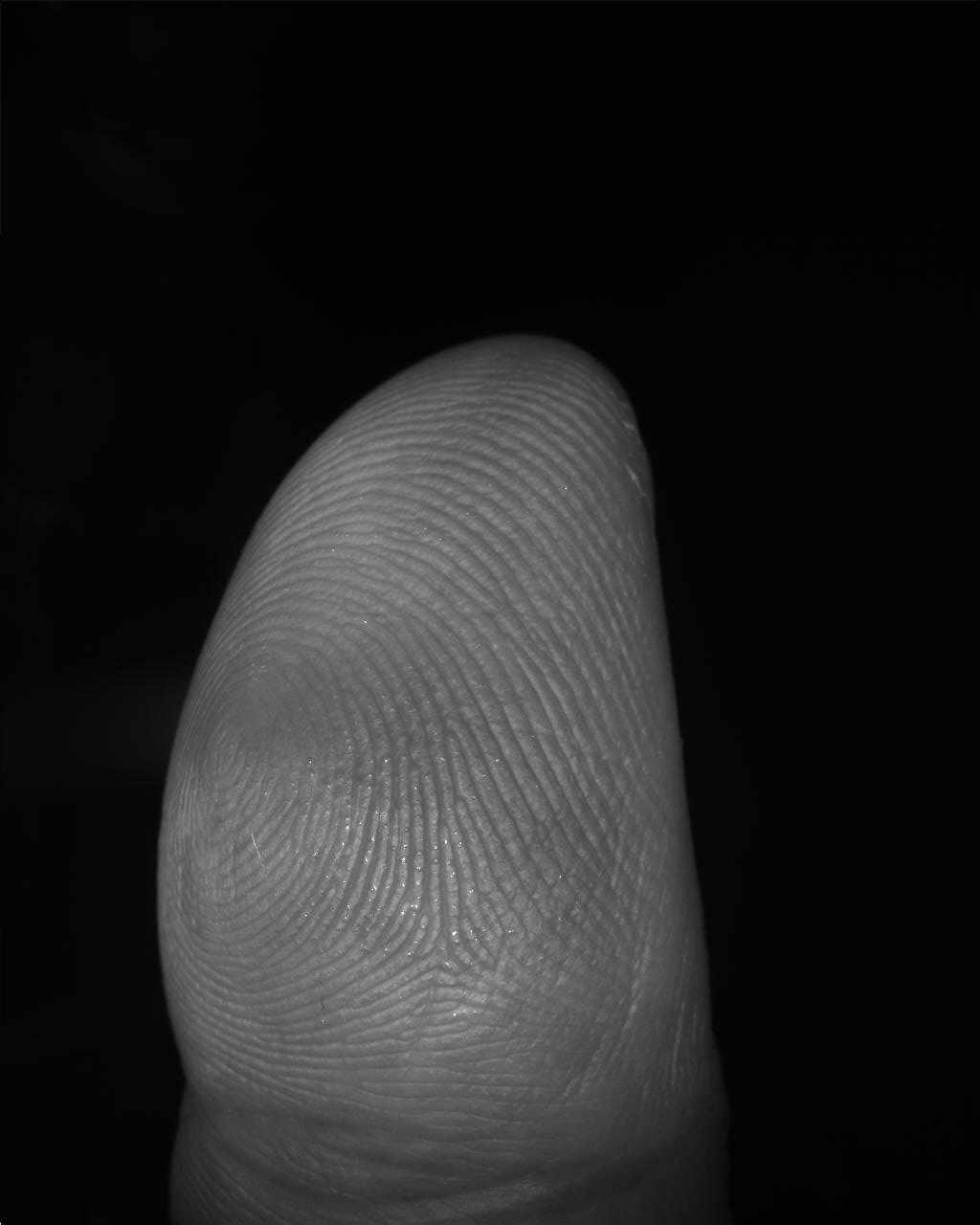}}\end{minipage} \\\begin{minipage}{0.1\textwidth}\centerline{\includegraphics[scale=0.08]{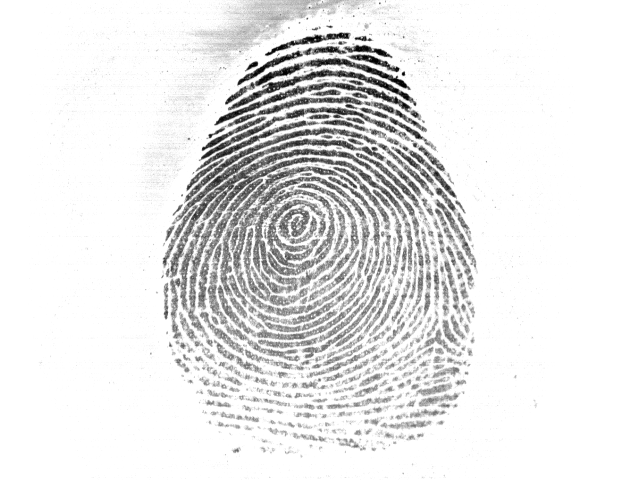}}\end{minipage}\begin{minipage}{0.1\textwidth}\centerline{\includegraphics[scale=0.2]{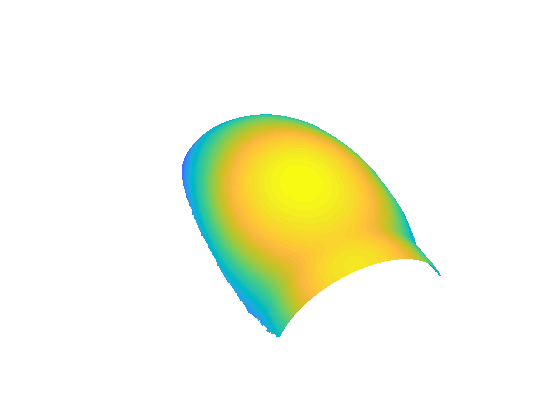}}\end{minipage}}&\tabincell{c}{\begin{minipage}{0.1\textwidth}\centerline{\includegraphics[scale=0.18]{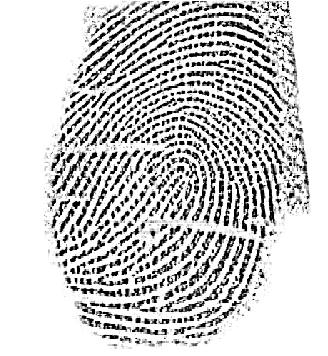}}\end{minipage}  \begin{minipage}{0.1\textwidth}\centerline{\includegraphics[scale=0.2]{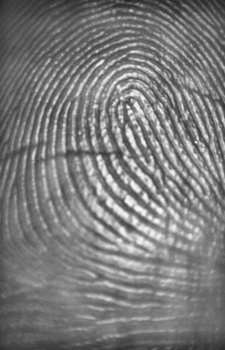}}\end{minipage}} \\
\hline
Image type & \tabincell{c}{Contactless 2D fingerprints\\ \& ground-truth depth data\\by structured light} & \tabincell{c}{Contactless 2D fingerprints\\ \& contact-based 2D fingerprints\\ \& 3D finger shapes reconstructed by\\our shape-from-silhouette algorithm}&\tabincell{c}{Contactless 2D fingerprints\\ \& contact-based 2D fingerprints}\\
\hline
Description & \tabincell{c}{220 contactless fingerprints\\ {\&} corresponding depth data\\acquired by structured light}&\tabincell{c}{9,000 contactless fingerprints\\from 1500 fingers $\times$ 3 angles $\times$ 2 captures\\{\&} mated 6,000 contact-based fingerprints\\from 1500 fingers $\times$ 4 captures}&\tabincell{c}{2,976 contactless fingerprints\\from 496 fingers $\times$ 6 captures\\{\&} mated 2,976 contact-based fingerprints\\from 496 fingers $\times$ 6 captures}\\
\hline
Training data &  \tabincell{c}{22,000 patches from 220 data,\\13,200 for training network} &\tabincell{c}{15,000 patches from the\\last 3,000 contactless fingerprints,\\9,000 patches for training network}&--\\
\hline
Experiment& Reconstruction accuracy &\tabincell{c}{Reconstruction accuracy\\{\&} Matching accuracy}&Matching accuracy\\
\hline
Experiment data&\tabincell{c}{8,800 patches for\\reconstruction experiment}&\tabincell{c}{6,000 patches for\\reconstruction experiment\\{\&} first 6,000 fingerprints for\\matching experiment}& \tabincell{c}{2,976 pairs of fingerprints for\\matching experiment}\\
\bottomrule
\end{tabular}
\label{table:database}
\end{table*}
To train the network, as shown in Table \ref{table:database}, we use two contactless fingerprint databases: The PolyU contactless 3D fingerprint database \cite{liu20143d}, and UNSW 2D/3D fingerprint database \cite{zhou2014benchmark}. This section introduces the usage of PolyU 3D database.

PolyU 3D database contains 220 contactless fingerprint images with corresponding ground truth depth maps obtained using the structured light technique. We preprocess the raw images using methods in Section \ref{ssec:preprocess}, then extract orientation and period features using VeriFinger \cite{verifinger}. The orientation and period extracted by VeriFinger are used as ground-truth. Although they may not be true values, our aim is to recover the depth map and computing orientation and period is just a reference here.

The original depth data of PolyU 3D database are very noisy and contain ridge-valley variability, which makes gradients very unstable. Therefore, we smooth the initial depth map using the moving least square \cite{fatehpuria2006acquiring}. We aim at estimating a smoothed finger shape without the 3D ridge-valley structure in this paper.

For each point $(x,y,z)$, we compute a quadric surface $f(x,y)$ using $K$ nearest neighbors:
\begin{equation}
\begin{aligned}
f(x,y)=a_1x^2+a_2xy+a_3y^2+a_4x+a_5y+a_6\\
\min_a\sum_{i=1}^Kw_i(z_i-f(x_i,y_i))^2,w_i=\exp\left(-\frac{(x_i-x)^2+(y_i-y)^2}{\epsilon^2}\right)
\end{aligned}
\label{equ:smooth}
\end{equation}

Solving equation (\ref{equ:smooth}) gets a smoothed estimation $f(x,y)$ to replace original depth $z$. Then the gradients are computed:
\begin{equation}
\begin{aligned}
(g_x,g_y)=\left(\frac{\partial f}{\partial x},\frac{\partial f}{\partial y}\right).
\end{aligned}
\label{equ:gradient}
\end{equation}
When resizing images in preprocessing, depth and gradient data are also scaled simultaneously. Then we crop training patches of size $512\times 512$ from image, mask, orientation, period, and gradient maps. For each pair of data in PolyU 3D database, 100 patches are extracted. Therefore, we get 22,000 sets of training data from PolyU 3D database. We use $60\%$ for training, and $40\%$ for testing.
\subsubsection{Synthetic Training Data}\label{sssec:data2}
Although PolyU 3D database contains ground-truth depth data, the contactless fingerprints in this database are all of front pose, lacking pose variations. Contactless fingerprints of side pose usually have steeper shapes and more severe perspective distortion. Also, the lighting condition of side pose may be different from front pose, which makes the image quality different. Therefore, side pose data are also needed to enlarge data comprehensiveness.

We use UNSW database \cite{zhou2014benchmark} to generate side pose data. The contactless fingerprints in this database are captured by Surround Imager \cite{parziale2006surround}. Images from three poses are captured simultaneously: left, middle, and right, and different poses are 45 degrees apart. Therefore, contactless fingerprints in this database have sufficient pose variations and can be used to train a network suitable for both front and side fingerprints.

\begin{figure}[!]
\centering
\centerline{\includegraphics[width=.95\linewidth]{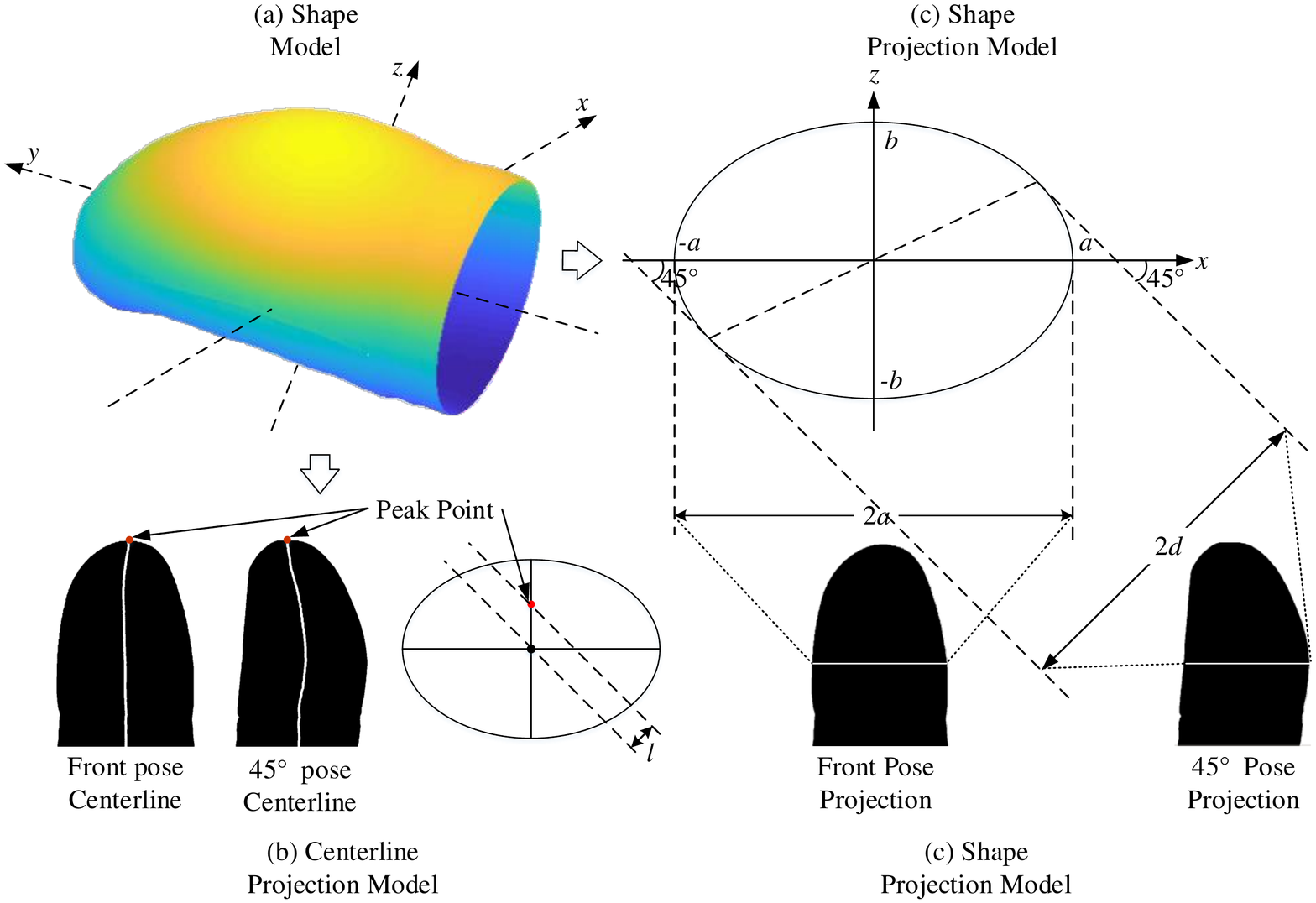}}
\caption{Finger shape projection model.}
\label{fig:shapemodel}
\end{figure}

Because this database has no ground-truth 3D fingerprint data, we design a shape from silhouette algorithm (\textbf{Algorithm \ref{alg:reconstruct_0}}) to generate finger shape models from images of three different angles. This dataset was captured by scanner of TBS Inc. \cite{parziale2006surround}, which uses shape-from-silhouette to reconstruct 3D finger shape. Therefore, shape-from-silhouette algorithm is suitable for UNSW database. And this algorithm is only used for getting a 3D finger shape from all three views, which is not the monocular fingerprint reconstruction algorithm in our manuscript.

Similar to \cite{tan2020towards}, a varying ellipse shape model is used to represent finger shape. Fig. \ref{fig:shapemodel} illustrates the projection model of different views, and we use a geometric method to compute ellipse parameters from different views of images. Based on the geometric relationship, we can get that:
\begin{equation}
\begin{aligned}
d=\sqrt{\frac{a^2+b^2}{2}},
\end{aligned}
\label{equ:shapemodel}
\end{equation}
where $a$ is the semi-major axis estimated from the front view, $d$ is the half projection length of the side view, and $b$ is the semi-minor axis to be estimated. 

The occluded area can be computed as:
\begin{equation}
\begin{aligned}
\theta=\arctan\frac{b^2}{a^2},
\end{aligned}
\label{equ:angle}
\end{equation}
where $\theta$ is the critical angle separating occluded area and non-occluded area. 

The pseudocode of shape from silhouette reconstruction method is given in \textbf{Algorithm \ref{alg:reconstruct_0}}. For a pair of contactless fingerprint images with 3 poses $I_{\text{F}}, I_\text{R}, I_\text{L}$, we first extract their contours using morphological operations. The contours are used to compute ellipse parameters $a, b, c_z$. 

Ellipse’s semi-major axis $a$ can be directly computed from front pose image $I_\text{F}$, as the front pose is the projection of the semi-major axis. Therefore, $a$ would be $d_\text{F}/2$, where $d_\text{F}$ is the projected length on front pose image $I_\text{F}$.

For semi-minor axis $b$, Equation (\ref{equ:shapemodel}) defines the relationship between $a, b$, and the side pose projected length $d$. Therefore, we use the right length $d_\text{R}$ to get a right pose estimation $b_\text{R}$, $d_\text{L}$ to get a left pose estimation $b_\text{L}$, and their average is the final estimation $b$.

Then we solve center translation $c_z$. The center points of different rows are the same on front pose image, but differ on side pose images. The spatial shift $l$ is caused by translation $c_z$, and $c_z=\sqrt{2}l$ according to projection model. Similar to estimating $b$, we estimate $c_z$ from left and right images, and use their average as the final result.

At last, we get a shape model, including an upper surface $z_\text{up}$, which is the front pose, and a lower surface $z_{\text{down}}$. Each corresponds to half of an ellipse. For side pose depth $z_\text{R}, z_\text{L}$, they are combinations of part of $z_{\text{up}}$ and $z_{\text{down}}$ because of rotation and occlusion. As shown in Fig. \ref{fig:shapemodel}, for right pose shape $z_\text{R}$, the piece of $[0, \theta]$ (start from the left endpoint of the ellipse) of $z_{\text{up}}$ is occluded. And we compensate $[0, \theta]$ (start from the right endpoint of the ellipse) of $z_{\text{down}}$ as the newly-seen area after rotation to get complete $z_\text{R}$. Noticing that current $z_\text{R}$ is still an observation by front view. Then it is rotated by $45^{\circ}$ to get the right view result. The left pose depth $z_\text{L}$ is symmetric.

\begin{algorithm}
\caption{Reconstruction Algorithm Using Multiview Images} 
\label{alg:reconstruct_0}
\hspace*{0.02in} {\bf Input:} 
Front, left, and right images $I_\text{F}, I_\text{R}, I_\text{L}$\\
\hspace*{0.02in} {\bf Output:} 
Depth $z_\text{F}, z_\text{R}, z_\text{L}$
\begin{algorithmic}[1]
\State Extract contours from $I_\text{F}, I_\text{R}, I_\text{L}$.
\For{each row $y$ } 
\State Computing line length $d_\text{F}, d_\text{R}, d_\text{L}$ from images
\State Ellipse semi-major axis $a=\frac{d_\text{F}}{2}$
\State Ellipse semi-minor axis $b$:
\State $b_\text{R}=\sqrt{\left(\frac{d_\text{R}}{2}\right)^2-a^2},b_\text{L}=\sqrt{\left(\frac{d_\text{L}}{2}\right)^2-a^2},b=\frac{b_\text{R}+b_\text{L}}{2}$ 
\State Computing center translation $l_\text{F},l_\text{R}$ from images
\State Center location $c_z=\frac{\sqrt{2}}{2}\left(l_\text{F}+l_\text{R}\right)$
\State Upper surface: $z_{\text{up}}(x,y)=\frac{b}{a}\sqrt{a^2-x^2}+c_z$
\State Lower surface: $z_{\text{down}}(x,y)=-\frac{b}{a}\sqrt{a^2-x^2}+c_z$
\State Computing vanishing angle $\theta$ from Equation \ref{equ:angle}
\State Stitching parts of $z_{\text{up}}$ and $z_{\text{down}}$ together to get $z_\text{R}$ and $z_\text{L}$
\EndFor
\State Rotating $z_\text{R}$ by $45^{\circ}$ around $y$ axis
\State Rotating $z_\text{L}$ by $-45^{\circ}$ around $y$ axis
\State $z_\text{F}=z_{\text{up}}$
\State \Return $z_\text{F}, z_\text{R}, z_\text{L}$
\end{algorithmic}
\end{algorithm}

UNSW database contains 9,000 images from 150 people $\times$ 10 fingers $\times$ 3 angles $\times$ 2 captures. Same as \cite{Dabouei2019}, we use the last 3,000 images from the last 50 people for training the network, others for the matching experiment. We use the abovementioned reconstruction method to simultaneously get shape models from three images of different poses. Therefore, we get $50\times 10 \times 3 \times 2=3000$ shape models. As we can get 3,000 data from UNSW database, which are much larger than the 220 data from PolyU 3D database, we only generate 5 patches from each data, and a total number of 15,000 patches of size $512\times 512$ are extracted for fine-tuning the network.
\subsection{Surface Reconstruction}\label{ssec:reconstruction}
We use the gradient output by the abovementioned network to reconstruct the surface. As gradients are derivatives of depths, depths can be recovered by integration. For reconstruction accuracy, we design a specific integration method (\textbf{Algorithm \ref{alg:reconstruct}}).

\begin{algorithm}
\caption{Surface Reconstruction Algorithm} 
\label{alg:reconstruct}
\hspace*{0.02in} {\bf Input:} 
Gradients $g_x,g_y$\\
\hspace*{0.02in} {\bf Output:} 
Depth $z$
\begin{algorithmic}[1]
\State Find integration starting point (zero point) $(c_x,c_y)$, which has the minimum gradients value.
\State Integration along $x-y$ path:
\For{$x$} 
\State $z_1(x,c_y)=\int_{c_x}^xg_x(x,c_y)dx$
\For{$y$} 
\State $z_1(x,y)=\int_{c_y}^yg_y(x,y)dy$
\EndFor
\EndFor
\State Integration along $y-x$ path:
\For{$y$} 
\State $z_2(c_x,y)=\int_{c_y}^yg_y(c_x,y)dy$
\For{$x$} 
\State $z_2(x,y)=\int_{c_x}^xg_x(x,y)dx$
\EndFor
\EndFor
\State $z(x,y)=\frac{z_1(x,y)+z_2(x,y)}{2}$
\State \Return $z$
\end{algorithmic}
\end{algorithm}

The basic idea of the reconstruction algorithm is to integrate outward from center to periphery. This is because the gradients in the center region are of small value, while those in the peripheral region are of large value. Integration from center to periphery could reduce numerical error. The integration starting point, or center point, is defined as the minimum gradient value.  

As line integral $\int g_x(x,y)dx+g_y(x,y)dy$ from $(0,0)$ to $(x,y)$ has infinite integral paths, we choose $(0,0)\rightarrow(x,0)\rightarrow(x,y)$ and $(0,0)\rightarrow(0,y)\rightarrow(x,y)$ as two paths to compute $z_1$ and $z_2$, and use their average as the final result. Trapezoidal integral is used for numerical calculation.

\textbf{Algorithm \ref{alg:reconstruct}} describes the whole reconstruction process. The estimated gradient $g_x, g_y$ is used for calculation. We first find the integration starting point (zero point) $(c_x,c_y)$ of the smallest gradients value, which is usually located at the center of the fingerprint. 

Then numerical integral is conducted. As mentioned above, two integral paths are used. Integral $z_1$ first goes along $x$ axis, then switch to $y$ axis. Integral $z_2$ is the opposite. The final result $z$ is their average to improve accuracy.
\subsection{Unwarping}\label{ssec:unwarping}
Contactless fingerprint unwarping simulates finger rolling action to unfold the finger surface into a flat plane. We develop an arc length based unwarping algorithm (\textbf{Algorithm \ref{alg:unwarp}}). The integration method is similar to the reconstruction algorithm, but replaces integrating gradient value with arc length. 

\begin{algorithm}
\caption{Contactless Fingerprint Unwarping Algorithm} 
\label{alg:unwarp}
\hspace*{0.02in} {\bf Input:} 
Gradients $g_x,g_y$, contactless image $I_{\text{in}}(x,y)$\\
\hspace*{0.02in} {\bf Output:} 
Unwarped image $I_{\text{out}}(x,y)$
\begin{algorithmic}[1]
\State Find integration starting point (zero point) $(c_x,c_y)$, which has the minimum gradients value.
\State Compute horizontal arc length $u$ and vertical arc length $v$:
\For{$x$} 
\State $u(x,y)=\int_{c_x}^x\sqrt{1+g_x(c_x,y)^2}dx$
\EndFor
\For{$y$} 
\State $v(x,y)=\int_{c_y}^y\sqrt{1+g_y(x,y)^2}dy$
\EndFor
\State Transform image $I_{\text{in}}$ to $I_{\text{out}}$ according to correspondences $(x,y)\rightarrow (u+c_x,v+c_y)$
\State \Return $I_{\text{out}}$
\end{algorithmic}
\end{algorithm}

The basic idea of the proposed unwarping algorithm is to use the arc length as a new coordinate. Therefore, the distance between two points on the surface is maintained after unwarping. We use the gradient to unwarp the image instead of using recovered depth to further reduce numerical error, as depth $z$ may already contains numerical error.

\textbf{Algorithm \ref{alg:unwarp}} gives the steps of the unwarping algorithm. First, we find the integration starting point the same as the reconstruction algorithm. Then curve distance is calculated using the basic curve length integration formula with gradients, and the arc length of each point is used as its new coordinate for image unwarping. A point at $(x,y)$ is moved to $(c_x+u,c_y+v)$ after transform. Bilinear interpolation is applied in image transform.
\subsection{Implementation Details} 
The proposed gradient estimation network is trained using 2 Nvidia 1080Ti GPUs using Adam \cite{adam2015} optimizer with default parameters. The losses are combinations of orientation, period, and gradient losses as described in Section \ref{sssec:losses} with parameters shown in Table \ref{table:para}. 

The network is first trained on PolyU 3D database, then fine-tuned on UNSW database. 13,200 samples from PolyU 3D and 9,000 samples from UNSW are used in training. As the network output has $1/8$ size of input size, the ground-truth orientation, period, and gradient are sampled to $64\times 64$, while the size of input image is $512\times 512$. The training batch-size is set to 16. The network converges after 50 epochs.

For testing, the output orientation, period, and gradient are up-sampled to the original size using bilinear interpolation. Estimation errors are computed between interpolated outputs and ground-truth. 8,800 samples from PolyU 3D and 6,000 samples from UNSW are used in testing. 

Table \ref{table:time} shows average time cost of each step of processing a single contactless fingerprint. The step of surface gradient estimation is implemented in Python and 2 Nvidia 1080Ti GPUs, while other steps are implemented in MATLAB and Intel Xeon E5-2640 2.5GHz CPU.
\begin{table}[!]
\caption{Average time cost of each step}
\centering
\begin{tabular}{cc}
\toprule
Step&Time cost (s)\\
\midrule
Image Preprocessing&0.97\\

Surface Gradient Estimation&0.46\\

Surface Reconstruction&0.68\\

Unwarping&1.07\\

Overall&3.18\\
\bottomrule
\end{tabular}
\label{table:time}
\end{table}
\section{Experiment}\label{sec:experiment}
This section includes experiments on the proposed reconstruction and unwarping algorithm. Section \ref{ssec:datasets}-\ref{ssec:matching_accuracy} are the introductions of the datasets used in experiments, the experiments on reconstruction accuracy, unwarping performances, and matching performances respectively. 
\subsection{Datasets}\label{ssec:datasets}
As summarized in Table \ref{table:database}, three databases are used for evaluating the proposed method. These databases are publicly available databases with ground-truth 3D depth data or strong view angle differences, which is suitable for training and testing the proposed network. Also, these databases are commonly used in the previous studies, and we can fairly compare our method with previous studies. 

\textbf{PolyU 3D database} \cite{liu20143d} contains contactless fingerprints with their ground-truth depth maps. This database is used for experiments for reconstruction accuracy. As there is only one image for one finger, matching experiments cannot be performed on this database. 

\textbf{UNSW database} \cite{zhou2014benchmark} contains contactless fingerprints and corresponding contact-based fingerprints. The ground-truth depth maps are generated using Algorithm \ref{alg:reconstruct_0} mentioned in Section \ref{sssec:data2}. Both reconstruction and matching experiments are performed on this database.

\textbf{PolyU CL2C database} \cite{lin2018matching} also contains contactless fingerprints and corresponding contact-based fingerprints. But as we cannot obtain the ground-truth 3D depth maps of this dataset, only matching experiments are conducted.
%
\subsection{3D Reconstruction Accuracy}\label{ssec:reconstruction_accuracy}
\subsubsection{Evaluation Protocols}
The proposed method is quantitatively evaluated on 8,800 samples from PolyU 3D database and 6,000 samples from UNSW database. Table \ref{table:error1} and \ref{table:error2} show quantitative results on mean errors of estimated orientation, period, gradient, and depth on these two databases. The definitions of four output errors are:
\begin{equation}
\begin{aligned}
e_{\text{Ori}}&=\frac{1}{|M|}\sum\min(|\text{Ori}_{\text{out}}-\text{Ori}_{\text{true}}|,\\
&180-|\text{Ori}_{\text{out}}-\text{Ori}_{\text{true}}|),\\
e_{\text{Ped}}&=\sqrt{\frac{1}{|M|}\sum(\text{Ped}_{\text{out}}-\text{Ped}_{\text{true}})^2},\\
e_{\text{Grad}}&=\sqrt{\frac{1}{|M|}\sum\|\text{Grad}_{\text{out}}-\text{Grad}_{\text{true}}\|_2^2},\\
e_{\text{Dep}}&=\sqrt{\frac{1}{|M|}\sum(\text{Dep}_{\text{out}}-\text{Dep}_{\text{true}})^2}.\\
\end{aligned}
\label{equ:err}
\end{equation}
$\text{Ori}_{\text{out}}$, $\text{Ped}_{\text{out}}$, $\text{Grad}_{\text{out}}$, and $\text{Dep}_{\text{out}}$ are the output orientation, period, gradient, and depth value. $\text{Ori}_{\text{true}}$, $\text{Ped}_{\text{true}}$, $\text{Grad}_{\text{true}}$, and $\text{Dep}_{\text{true}}$ are the ground-truth orientation, period, gradient, and depth value. $1/|M|$ means averaging within the mask region.

The average error means the average value of Equation (\ref{equ:err}) among all test samples. The average weighted error means adding weight using Equation (\ref{equ:weight}) when computing Equation (\ref{equ:err}) to balance between the center area and edge area. There are several notes for evaluations:
\begin{itemize}
\item The orientation and period maps are just intermediate outputs to examine whether the network properly extracts fingerprint texture features. Gradient output is the final required result, while orientation and period outputs are just referencing here.
\item The gradient and depth values from PolyU 3D database are acquired by structured light device. Therefore, they are sufficiently accurate as ground-truth, but contain only front pose data. Meanwhile, the depth data from UNSW database are synthetic data, as mentioned in Section \ref{sssec:data2}. But they contain both front pose and side pose data.
\item The units of errors of four outputs are different: the orientation error is in degree; the period error is in pixel; the depth error is in mm; and the gradient error is in mm/pixel. 
\item The estimated depth map has scale uncertainty. The reconstructed depth from Section \ref{ssec:reconstruction} is the integration result from zero point, thus the minimum value in the finger surface is always zero. Therefore, ground-truth depth map is subtracted a number to make the depth value of the corresponding lowest point is zero. Then, depth error is computed between them.
\end{itemize}

\begin{table}[!]
\caption{Evaluation results on PolyU 3D database}
\centering
\begin{tabular}{ccccc}
\toprule
Errors&\tabincell{c}{Orientation\\(Degree)}&\tabincell{c}{Period\\(Pixel)}&\tabincell{c}{Gradient\\(mm/Pixel)}&\tabincell{c}{Depth\\(mm)}\\
\midrule
Average&2.6927&0.2624&0.4359&0.1802\\

\tabincell{c}{Weighted}&2.8698&0.2722&0.0035&0.1506\\
\bottomrule
\end{tabular}
\label{table:error1}
\end{table}
\begin{table}[!]
\caption{Evaluation results on UNSW database}
\centering
\begin{threeparttable}
\begin{tabular}{ccccc}
\toprule
Errors&\tabincell{c}{Orientation\\(Degree)}&\tabincell{c}{Period\\(Pixel)}&\tabincell{c}{Gradient\\(mm/Pixel)}&\tabincell{c}{Depth\\(mm)}\\
\midrule
\tabincell{c}{All$^1$}&2.7056&0.3376&0.1599&0.4844\\

\tabincell{c}{All$^2$}&2.5170&0.3287&0.0073&0.3540\\

\tabincell{c}{Front Pose$^1$}&2.5297&0.3355&0.0679&0.3236\\

\tabincell{c}{Front Pose$^2$}&2.6690&0.3471&0.0046&0.1626\\

\tabincell{c}{Side Pose$^1$}&2.7935&0.3386&0.2059&0.5648\\

\tabincell{c}{Side Pose$^2$}&2.4410&0.3194&0.0086&0.4512\\
\bottomrule
\end{tabular}
\begin{tablenotes}
\item 1: Average error.
\item 2: Weighted average error.
\end{tablenotes}
\end{threeparttable}
\label{table:error2}
\end{table}
\subsubsection{Evaluation Results}
\begin{figure}[!]
\centering
\centerline{\includegraphics[width=\linewidth]{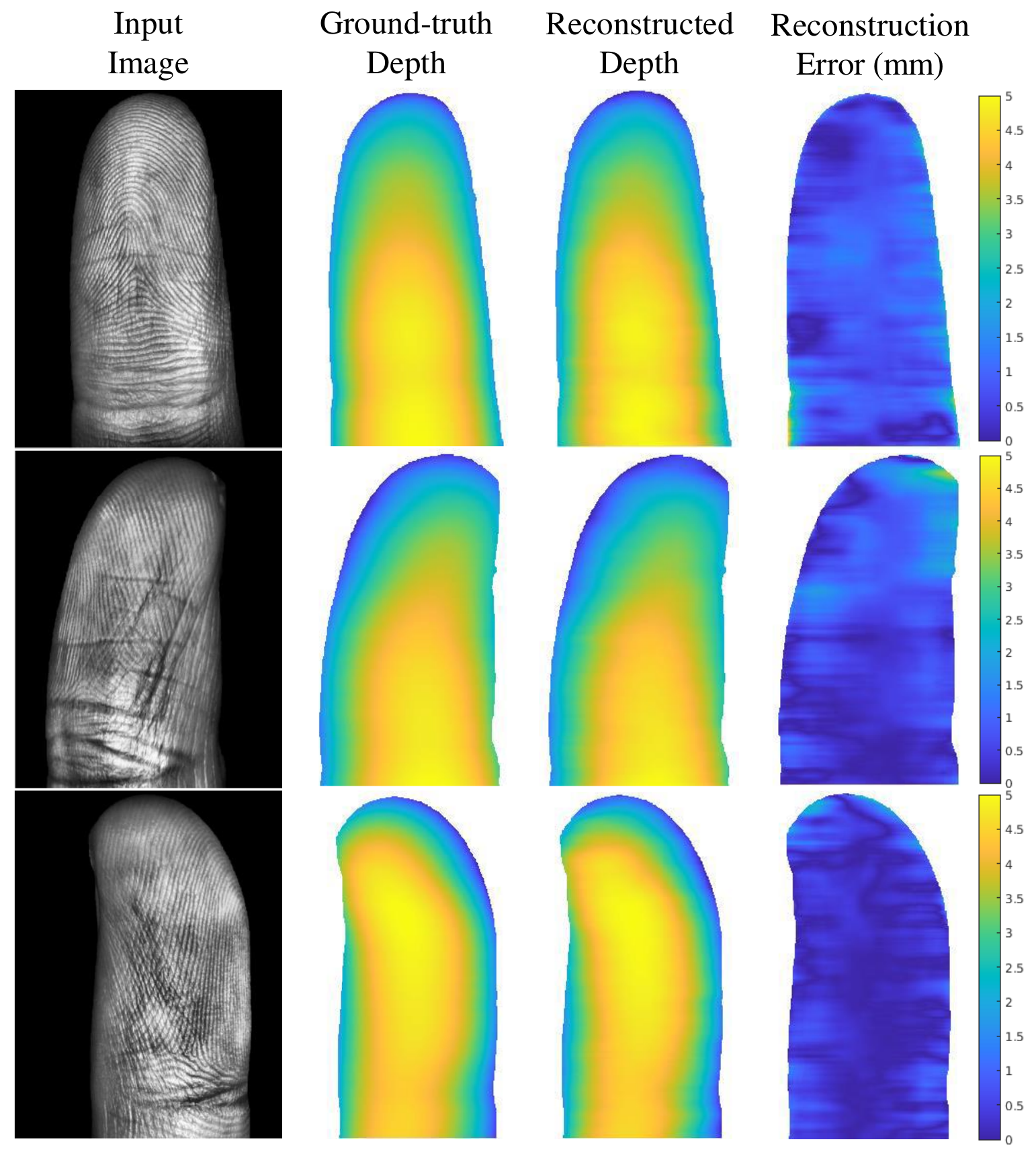}}
\caption{Illustration of reconstructed 3D finger shape and reconstruction error. Most reconstruction errors occur near the border region.}
\label{fig:reconstruct_error}
\end{figure}

Table \ref{table:error1} and \ref{table:error2} show evaluation results on the Poly3D database and UNSW database. From Table \ref{table:error1} and \ref{table:error2}, we can see that:
\begin{itemize}
\item Adding weights barely makes a difference in orientation and period estimation errors. But gradient and depth errors are reduced after adding weights, especially for gradients. As Fig. \ref{fig:reconstruct_error} shows, most depth errors occur near the edge region, while the center region is of small error and is more reliable. This is also a motivation for designing of our reconstruction and unwarping algorithm that the calculation should start from the center area to reduce error.
\item A large part of side pose fingerprints have larger gradient and depth errors than front pose fingerprints. This is because side pose images are usually more curved. Therefore, they are more challenging to estimate shapes and have larger errors.
\end{itemize}

\begin{table}[!]
\caption{Numerical errors of reconstruction algorithm on PolyU 3D database}
\centering
\begin{tabular}{cc}
\toprule
\tabincell{c}{Reconstruction Error}&\tabincell{c}{Depth (mm)}\\
\midrule
\tabincell{c}{Complete Region}&2.5210\\

\tabincell{c}{Region without Margin}&0.0141\\
\bottomrule
\end{tabular}
\label{table:algorithm_error}
\end{table}

To further study the accuracy of the proposed reconstruction algorithm, we conduct a numerical experiment. Table \ref{table:algorithm_error} shows the numerical errors of the reconstruction algorithm on PolyU 3D database. Here numerical errors refer to using the ground-truth gradient to reconstruct the depth, and computing the differences between ground-truth depth and reconstructed depth. This is an evaluation of the numerical precision of our reconstruction algorithm, as both input and target are ground-truth data. The complete region in Table \ref{table:algorithm_error} means computing RMSE within the whole fingerprint region, while the region without margin means removing three pixels from the fingerprint border. 

Clearly, from Table \ref{table:algorithm_error}, nearly all of the numerical errors happen in the border region. This is because numerical differential and integration are used in our algorithm, and they are unstable near the border region as sudden changes of depth happen here. Removing 3-pixel wide border barely affects the final result. In our following experiments, all results are computed without border region.

We can see from Table \ref{table:algorithm_error} that the numerical error is relatively small ($\sim0.01$mm) compared with the reconstruction error in the following experiment section ($\sim0.2$mm). Therefore, the numerical error of the reconstruction algorithm is neglected during the following reconstruction experiment. The reconstruction errors mainly come from gradient estimation errors.

Fig. \ref{fig:reconstruct_error} shows some qualitative results of reconstruction errors. Three examples of front, right, and left contactless fingerprints are displayed. We can see from heatmaps of reconstruction errors that most errors are relatively small ($<0.5$mm). Large reconstruction errors ($>2$mm) happen mostly in the border region due to numerical integration errors or foreground segmentation errors. The mean reconstruction errors are very small if we remove these border outliers, which is consistent with the evaluation results in Table \ref{table:error1}, \ref{table:error2}, and \ref{table:algorithm_error}.

\begin{figure}[!]
\centering
\centerline{\includegraphics[width=\linewidth]{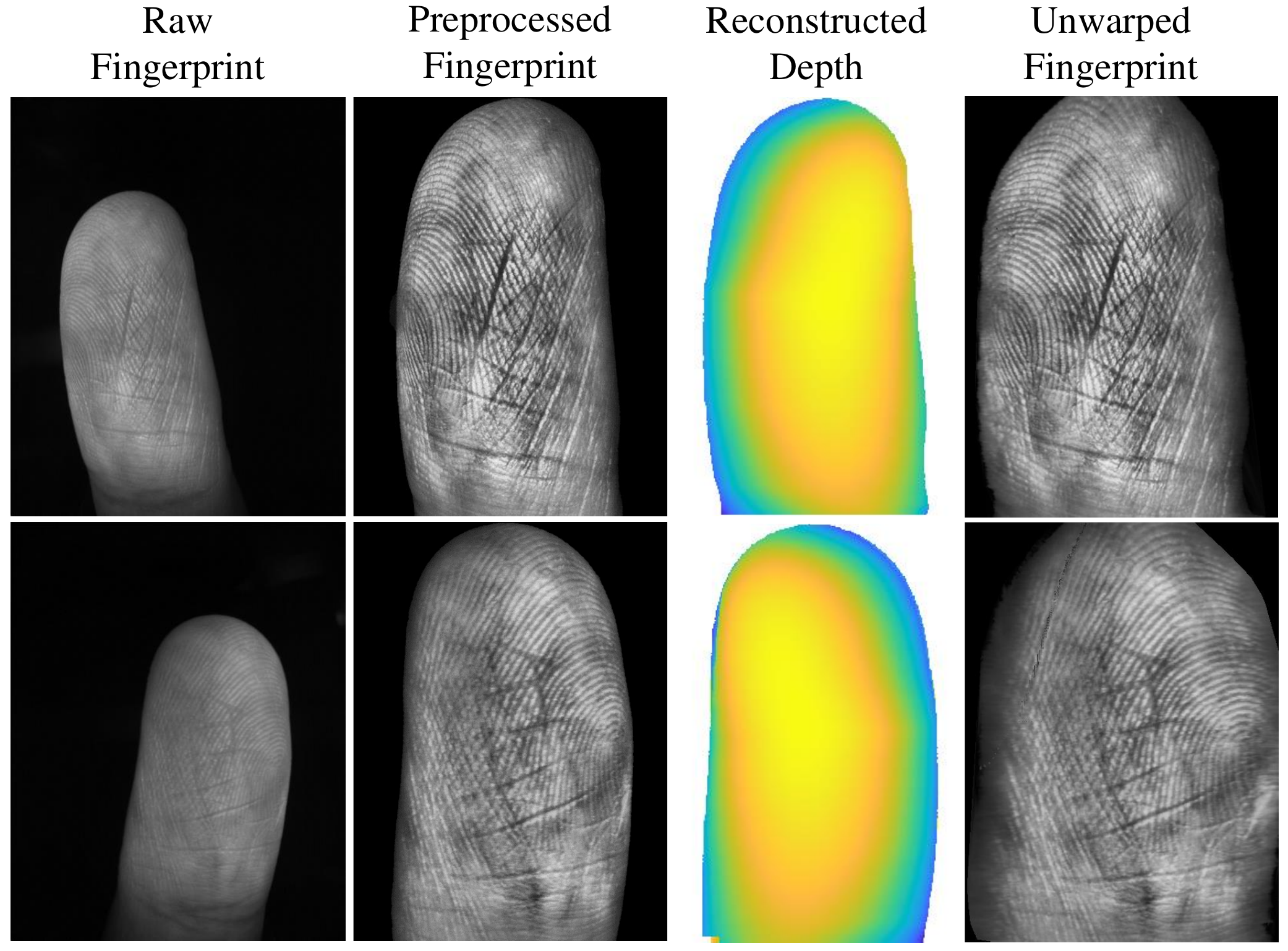}}
\caption{Illustrations of reconstruction and unwarping results of different poses from the same finger.}
\label{fig:yaw_example}
\end{figure}
Fig .\ref{fig:yaw_example} shows an example of the reconstruction and unwarping results by different poses from the same finger. Our method can successfully process fingerprints that varying in yaw angle, which is benefited from the preprocessing step that corrects fingerprint to the same pose.
\subsubsection{Ablation Study of Network Structure}\label{sssec:discussion_4}
The proposed network learns gradient, orientation field, and ridge period simultaneously. But the orientation field and ridge period are only involved in the training process, and are not utilized in testing. The orientation field and ridge period serve as auxiliary tasks here to assist learning the gradient, which is the main advantage of the multi-task learning structure \cite{Ruder2017overview}.

\begin{figure*}[!]
\centering
\centerline{\includegraphics[width=\linewidth]{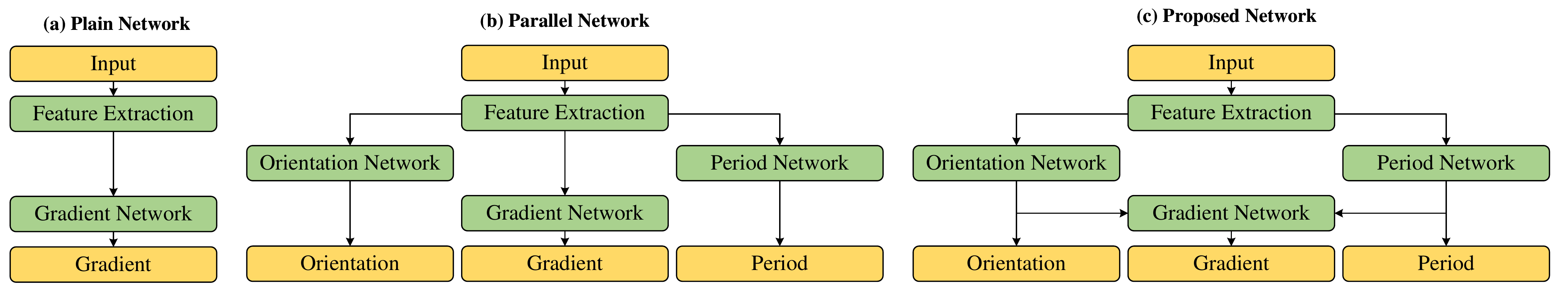}}
\caption{Illustrations of different network structures.}
\label{fig:network_ablation}
\end{figure*}

We conduct an ablation test on the network structure to examine the advantage of jointly learning orientation field and ridge period with gradient. As shown in Fig. \ref{fig:network_ablation}, three network structures are compared. Plain network means the proposed network without the orientation and period branches, only the gradient estimation network is utilized. Parallel network means estimating ridge orientation, period, and gradient parallelly, while the three branches in our network are connected. Fig. \ref{fig:network_ablation} shows the skeletons of different networks. The three networks are all trained on the training set described in section \ref{ssec:datasets}, and tested on PolyU 3D database. The three networks are trained with the same batch size and epochs.

Table \ref{table:eer_network} shows reconstruction errors on PolyU 3D database by different network structures. Our network has better gradient estimation and depth reconstruction results, which proves that using orientation and period is beneficial for estimating gradient. This is probably because the changes in ridge orientation and period directly reflect the finger surface inclination based on shape from texture theory \cite{blostein1989shape}. Estimating ridge orientation and period can be viewed as pre-tasks for estimating gradients. They also serve as middle layer outputs to judge if our network extracts meaningful fingerprint features.

\begin{table}[!]
\caption{Reconstruction errors on PolyU 3D database by different network structures}
\centering
\begin{tabular}{cccc}
\toprule
Errors&\tabincell{c}{Plain\\Network}&\tabincell{c}{Parallel\\Network}&\tabincell{c}{Our\\Network}\\
\midrule
\tabincell{c}{Gradient (mm/Pixel)}&0.7054&0.4446&0.4359\\

\tabincell{c}{Depth (mm)}&0.3502&0.2607&0.1802\\
\bottomrule
\end{tabular}
\label{table:eer_network}
\end{table}
\subsection{Unwarping Performance}\label{ssec:unwarping_accuracy}
\subsubsection{Unwarping Influences on Minutiae Matching}\label{sssec:discussion_3}
\begin{figure*}[!]
\begin{minipage}[b]{.45\linewidth}
\centering
\centerline{\includegraphics[width=\linewidth]{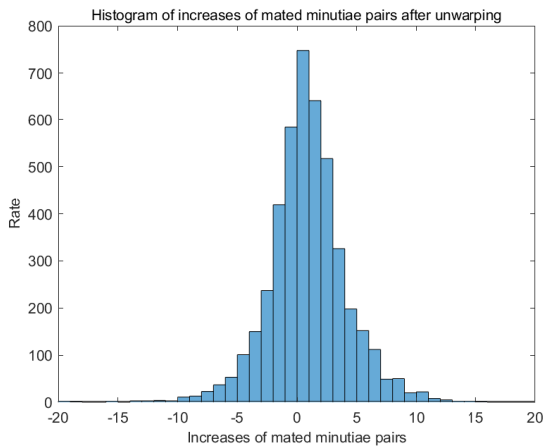}}
\centerline{(a) Contactless-contact matching}
\end{minipage}
\hfill
\begin{minipage}[b]{.45\linewidth}
\centering
\centerline{\includegraphics[width=\linewidth]{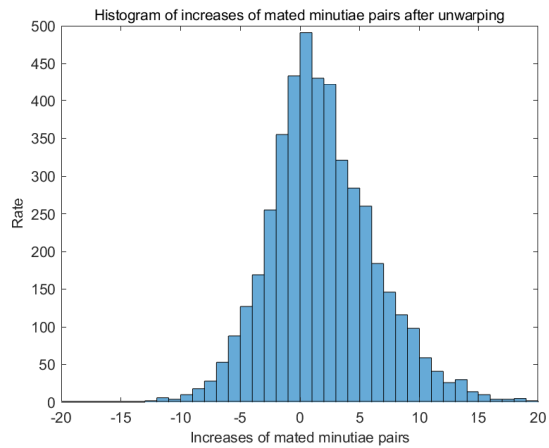}}
\centerline{(b) Contactless-contactless matching}
\end{minipage}
\caption{Histogram of increases of mated minutiae pairs after unwarping on UNSW database.}
\label{fig:minu}
\end{figure*}

\begin{figure*}[!]
\begin{minipage}[b]{.45\linewidth}
\centering
\centerline{\includegraphics[width=\linewidth]{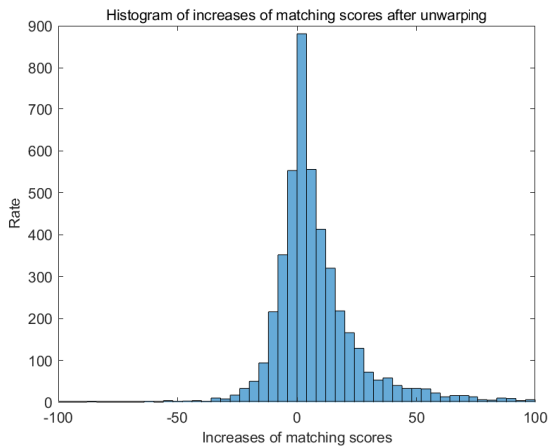}}
\centerline{(a) Contactless-contact matching}
\end{minipage}
\hfill
\begin{minipage}[b]{.45\linewidth}
\centering
\centerline{\includegraphics[width=\linewidth]{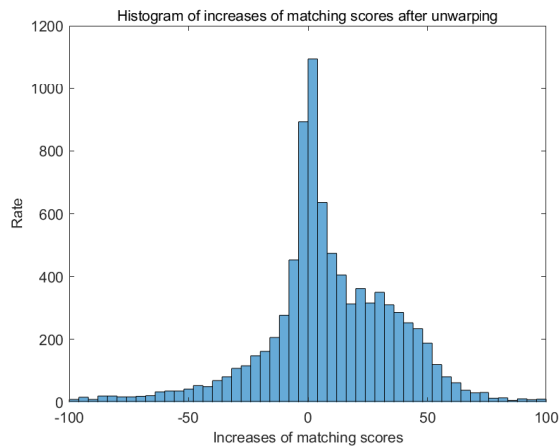}}
\centerline{(b) Contactless-contactless matching}
\end{minipage}
\caption{Histogram of increases of matching scores after unwarping on UNSW database.}
\label{fig:score}
\end{figure*}
We study the impact of the proposed unwarping algorithm on contactless fingerprint matching. Fig. \ref{fig:minu} is the histogram of increases of mated minutiae pairs after unwarping on UNSW database. For a pair of mated contactless/contact fingerprints, minutiae pairs are computed by VeriFinger. It can be seen from Fig. \ref{fig:minu} that the numbers of minutiae pairs increase after unwarping,

Fig. \ref{fig:score} is the histogram of increases of matching scores after unwarping by VeriFinger on UNSW database. And Fig. \ref{fig:score} shows that the matching scores of corresponding contactless{/}contact fingerprints increase after unwarping. Therefore, we can conclude that the proposed unwarping algorithm improves matching performances of contactless fingerprint matching.
\subsubsection{Comparison of Unwarping Algorithms}\label{sssec:discussion_2}
We conduct comparison experiments of three unwarping methods are compared: 1) non-parametric unwarping, 2) cylindrical unwarping (our method only horizontally), and 3) ellipsoidal unwarping (our method). 

Non-parametric unwarping \cite{fatehpuria2006acquiring}\cite{Shafaei2009new} keeps local structures by minimizing distance changes after unwarping:
\begin{equation}
\begin{aligned}
e=\sum_{j\in \mathcal{N}_i}\left(d_{ij}-r_{ij}\right)^2,
\end{aligned}
\label{equ:unwarp_energy}
\end{equation} 
where point $j$ is in the neighborhood of point $i$, $d_{ij}$ is the current distance between $i$ and $j$ after unwarping in 2D image space, and $r_{ij}$ is the original distance in 3D space. Minimizing the energy function in Equation (\ref{equ:unwarp_energy}) requires multiple iterations to reach the equilibrium state. Thus, this method is time-consuming.

Cylindrical unwarping here refers to unwarping fingerprint only horizontally like rolling a cylinder, while ellipsoidal unwarping (our method) unwarps in both directions like rolling an ellipsoid. It needs to be noticed that the cylindrical and ellipsoidal unwarping here are different from previous unwarping studies \cite{chen2006touchless}\cite{Sollinger2021}. In previous studies, a cylinder or ellipse/ellipsoid geometric model is used in unwarping, while our method uses gradient value to compute geodesic distance on the surface. Therefore, our method does not rely on any prior shape model, and is more universal. 

\begin{figure}[!]
\centering
\centerline{\includegraphics[width=\linewidth]{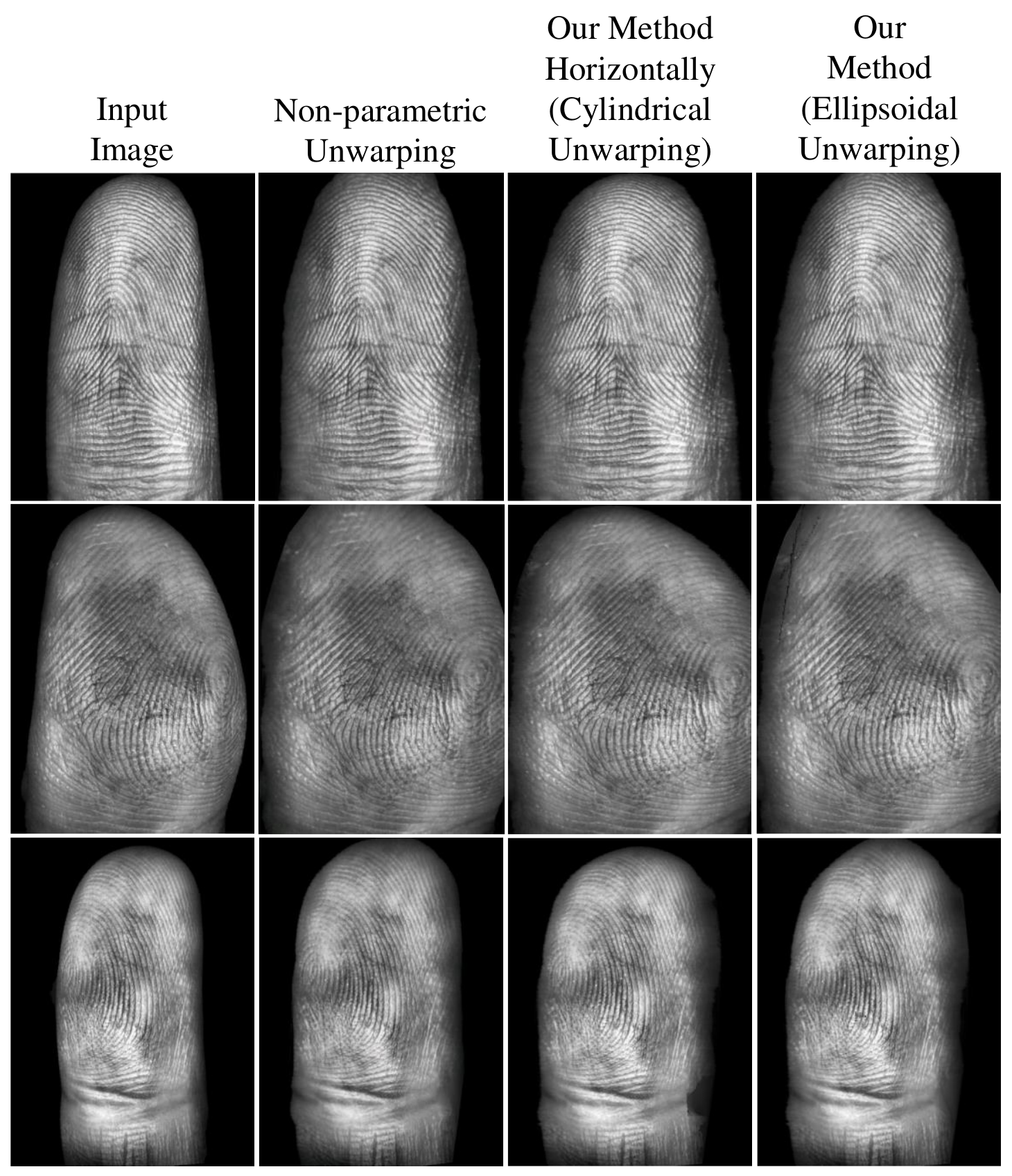}}
\caption{Examples of different unwarping algorithms.}
\label{fig:unwarping_algorithms}
\end{figure}

Figure \ref{fig:unwarping_algorithms} shows the results of different unwarping algorithms. All three methods can get reasonable unwarping results. Non-parametric unwarping is based on local structures, hard to  preserve long-term information. And its time cost is very high due to iterations. Cylindrical unwarping unwarps only horizontally, therefore only horizontal distortion is solved. Our method unwarps in both directions, and the perspective distortion along both axes is handled.

\begin{table}[!]
\caption{Equal error rates of matching results on UNSW database by different unwarping algorithms}
\centering
\resizebox{\linewidth}{!}{
\begin{tabular}{ccccc}
\toprule
\tabincell{c}{Matching\\Experiment}&\tabincell{c}{No\\Unwarping}&\tabincell{c}{Non-para\\Unwarping}&\tabincell{c}{Our\\Method\\Horizontally}&\tabincell{c}{Our\\Method}\\
\midrule
\tabincell{c}{CL-C}&15.79\%&14.43\%&16.28\%&{14.03}\%\\

\tabincell{c}{CL-CL}&23.95\%&20.53\%&22.37\%&{20.18}\%\\

\tabincell{c}{Time cost (s)}&--&140&1.06&1.07\\
\bottomrule
\end{tabular}
}
\label{table:eer_unwarp}
\end{table}
\subsection{Matching Accuracy}\label{ssec:matching_accuracy}
\subsubsection{Matching Protocols}
Matching experiments are performed on UNSW database and PolyU CL2C database. Two types of matching experiments are conducted: contactless to contact-based matching, and contactless to contactless matching. The matching scores are generated using VeriFinger 12.0 \cite{verifinger}.

\textbf{UNSW database.} Matching experiments are performed on the first 1,000 fingers of UNSW database. For contactless-contact matching, as there are two captures for each contactless fingerprint and four captures for each contact-based fingerprint, we use the first capture of contactless and contact fingerprint for efficiency. Therefore, a total number of $3,000\times 1,000$ pairs of matching is performed. As the 3,000 contactless fingerprints are from three different poses, we further divide the contactless-contact matching into three groups according to pose: front, right, and left. Each group contains $1,000\times 1,000$ pairs, where only 1,000 pairs are genuine matching, and the rest pairs are impostor matching. 

For contactless-contactless matching, we use the first capture of each contactless fingerprint as the reference fingerprint, and the second capture as the input fingerprint. Therefore, $3,000\times 3,000$ pairs of matching are performed. Noticing that matching between fingerprints of different poses from the same finger is also considered as genuine matching, there are 9,000 pairs of genuine matching in total, and the rest are impostor matching. 

In both matching experiments, the abovementioned matching is conducted on four types of images: raw images, unwarped images, rotated images, and rotated+unwarped images. The rotation here refers to pose compensation in \cite{labati2013contactless} and \cite{tan2020towards} that rotates contactless fingerprints of different poses to the same viewing angle. In our experiment, after estimating the 3D shape from the contactless fingerprint, we rotate all 3D shapes to the same front pose, then re-project the 3D finger to the 2D plane to get a rotated image. 

\begin{figure}[!]
\centering
\centerline{\includegraphics[width=\linewidth]{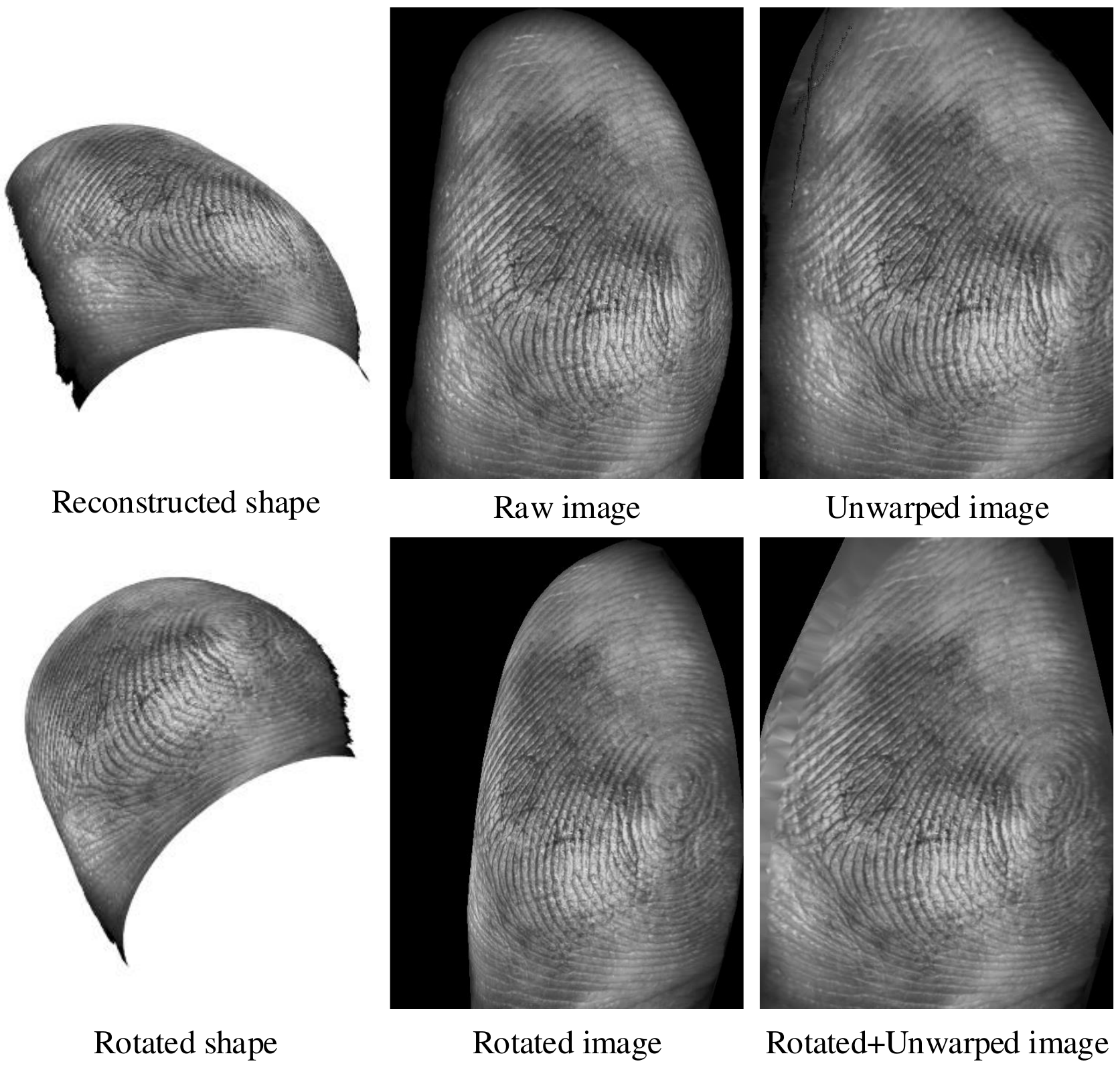}}
\caption{Illustration of fingerprint unwarping and rotation in matching experiment. The first row is the reconstructed 3D shape, raw image, and unwarped image. The second row is the rotated 3D shape, rotated image, and rotated+unwarped image.}
\label{fig:instance}
\end{figure}

The contactless fingerprints in UNSW database are captured by Surround Imager \cite{parziale2006surround} cameras placed $45^{\circ}$ apart. Therefore, the rotation angle is considered a known value of $\pm 45^{\circ}$, and is used in rotation compensation. Noticing that for front pose fingerprints, their rotated images are the same as raw images. Fig. \ref{fig:instance} shows the four images of unwarped results of a side pose fingerprint in the matching experiments.

\textbf{PolyU CL2C database.} PolyU CL2C database contains 2,976 contactless fingerprints and corresponding contact-based fingerprints from 496 fingers $\times$ 6 captures. For contactless-contact matching, there are $496\times6\times6=17,856$ genuine matchings and $496\times6\times495\times6=8,838,720$ impostor matchings. For contactless-contactless matching, there are $496\times15=7,440$ genuine matchings and $2976\times2975/2-7440=4,419,360$ impostor matchings. 

As there are no pose differences in contactless fingerprints of PolyU CL2C database, matching experiments are run only on raw images and unwarped images.

\begin{figure*}[htb]
\begin{minipage}[b]{.45\linewidth}
\centering
  \centerline{\includegraphics[width=\linewidth]{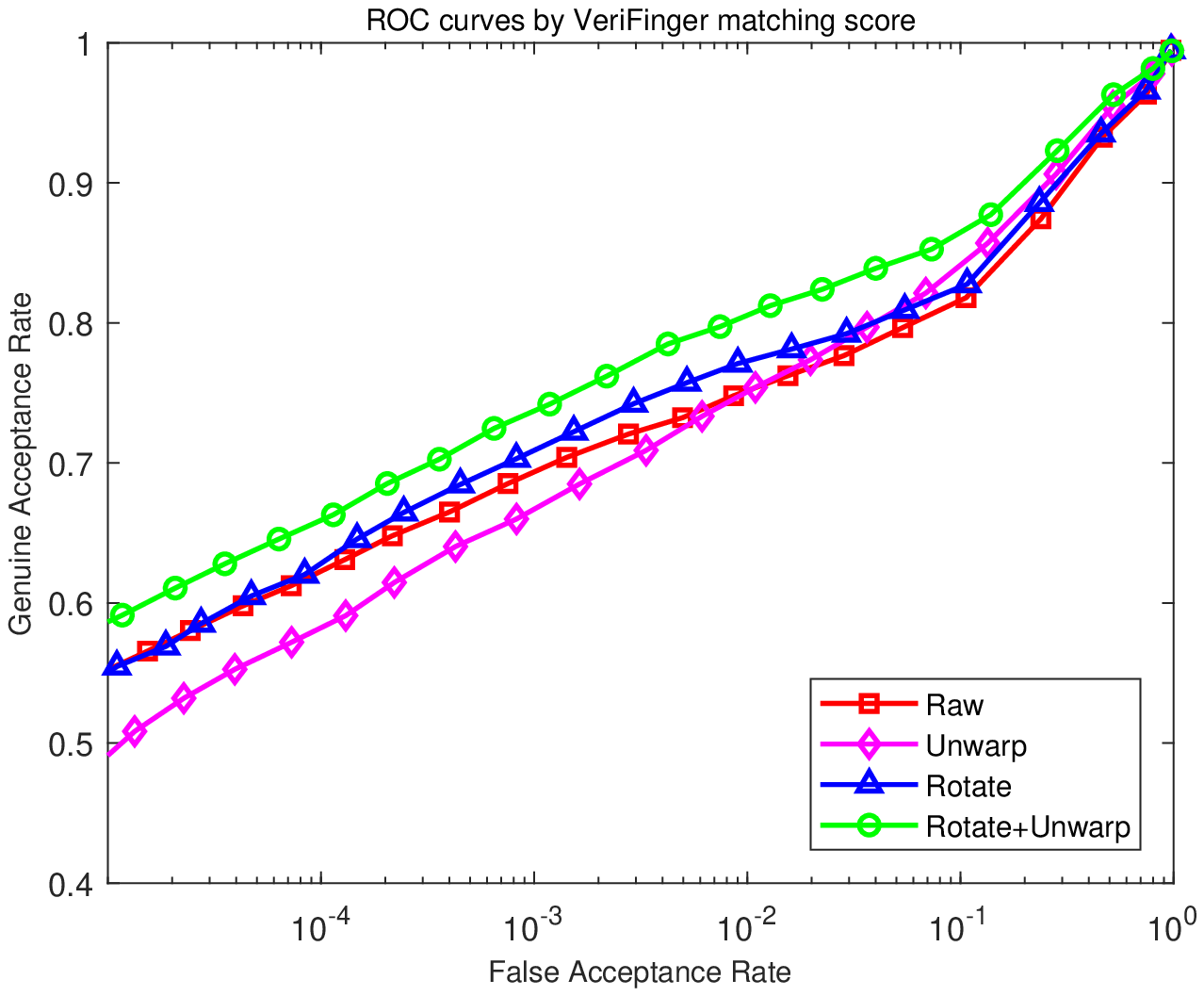}}
\centerline{(a) UNSW database}
\end{minipage}
\hfill
\begin{minipage}[b]{.45\linewidth}
\centering
  \centerline{\includegraphics[width=\linewidth]{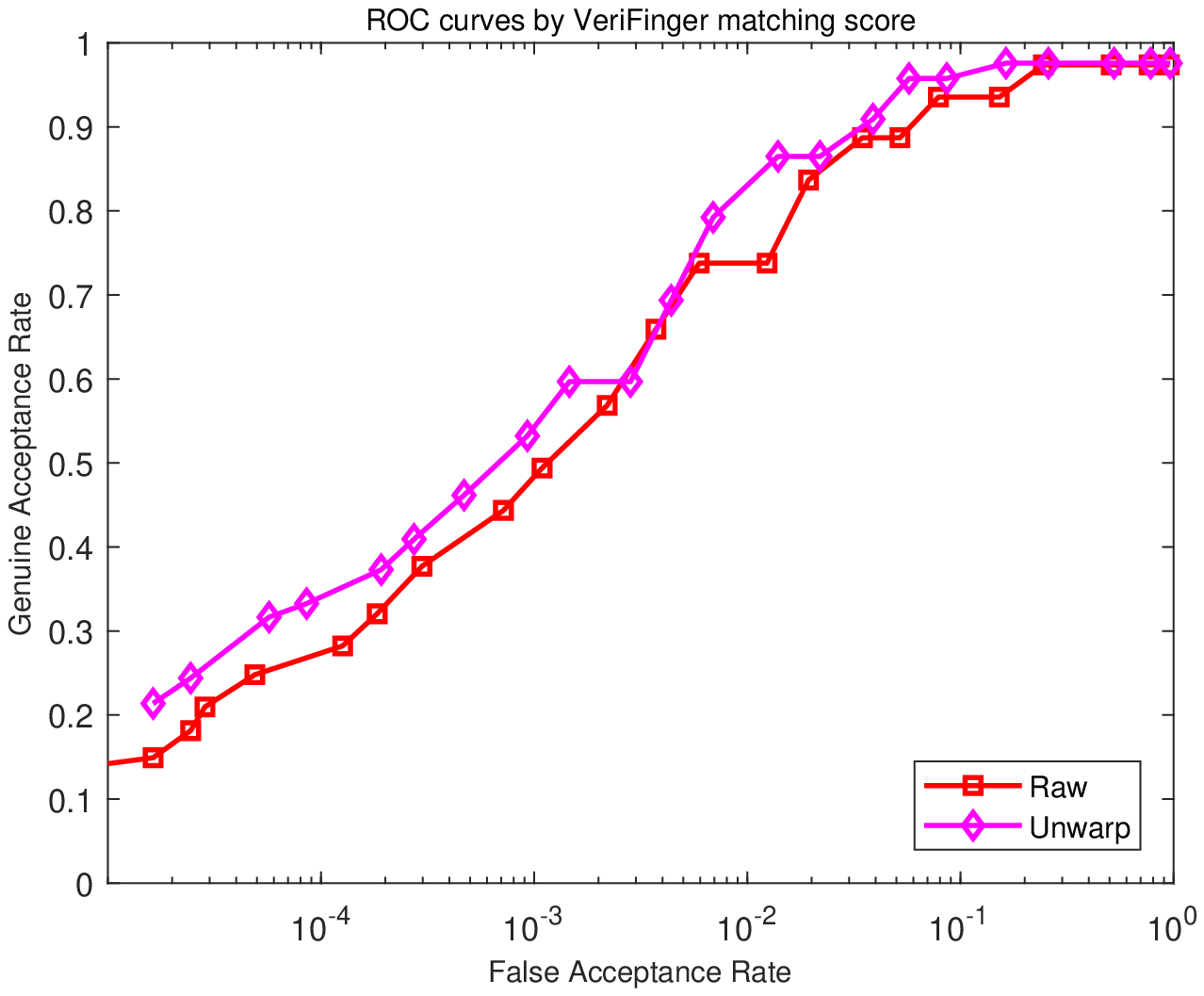}}
\centerline{(b) PolyU CL2C database}
\end{minipage}
\caption{ROC curves of contactless-contact fingerprint matching on two databases.}
\label{fig:cmc1}
\end{figure*}
\subsubsection{Contactless-Contact Matching}
Fig. \ref{fig:cmc1} shows the ROC curves of contactless-contact fingerprint matching on UNSW and PolyU CL2C databases. The proposed unwarping algorithm successfully improves the matching accuracy of contactless fingerprints, which proves that our unwarping algorithm is effective for reducing perspective distortion and improving matching performance. Table \ref{table:eer1} shows the detailed matching accuracy of contactless fingerprints with different poses on UNSW database, and Table \ref{table:eer11} shows the results on PolyU CL2C database. The unwarping method improves the matching accuracy of all poses on UNSW database. 
\begin{table}[!]
\caption{Equal error rates of contactless-contact matching on UNSW database}
\centering
\begin{tabular}{ccccc}
\toprule
Pose& Raw&Unwarp&Rotate&\tabincell{c}{Rotate\\+Unwarp}\\
\midrule
Front	&4.00\%	&\textbf{3.59}\%	&4.00\%	&\textbf{3.59}\%\\

Right	&19.31\%	&17.07\%	&17.84\%	&\textbf{16.47}\%\\

Left	&21.67\%	&18.70\%	&20.31\%	&\textbf{16.01}\%\\

All		&15.79\%	&14.03\%	&14.98\%	&\textbf{12.72}\%\\
\bottomrule
\end{tabular}
\label{table:eer1}
\end{table}
\begin{table}[!]
\caption{Equal error rates of matching results on PolyU CL2C database}
\centering
\begin{tabular}{ccc}
\toprule
\tabincell{c}{Matching\\Experiment}& Raw&Unwarp\\
\midrule
\tabincell{c}{Contactless-Contact}&6.45\%	&\textbf{5.73}\%\\

\tabincell{c}{Contactless-Contactless}&8.43\%	&\textbf{6.45}\%\\
\bottomrule
\end{tabular}
\label{table:eer11}
\end{table}

We compare our result with \cite{Dabouei2019}, which reported state-of-the-art performance on UNSW database. In \cite{Dabouei2019}, only front pose fingerprints are used in the matching experiment, which are much easier to match than fingerprints of other poses. Therefore, we compare \cite{Dabouei2019} with our front pose matching result. Meanwhile, we compare with \cite{lin2019cnn} on PolyU CL2C database. As Table \ref{table:acc1} shows, our method is superior to previous methods on contactless-contact matching on both databases. 

\begin{table}[!]
\caption{Matching performances of the proposed method}
\centering
\begin{tabular}{cccc}
\toprule
Database&\tabincell{c}{Matching\\Experiment}&\tabincell{c}{Our Method\\EER(\%)}&\tabincell{c}{Previous SOTA\\EER(\%)}\\
\midrule
UNSW &\tabincell{c}{CL-C}&3.59&7.71\cite{Dabouei2019} \\

PolyU CL2C&\tabincell{c}{CL-C}&5.73&7.93\cite{lin2019cnn} \\

UNSW &\tabincell{c}{CL-CL}&14.12&14.27\cite{tan2020towards} \\

PolyU CL2C&\tabincell{c}{CL-CL}&6.45&-- \\
\bottomrule
\end{tabular}
\label{table:acc1}
\end{table}

\subsubsection{Contactless-Contactless Matching}
\begin{figure*}[!]
\begin{minipage}[b]{.45\linewidth}
\centering
\centerline{\includegraphics[width=\linewidth]{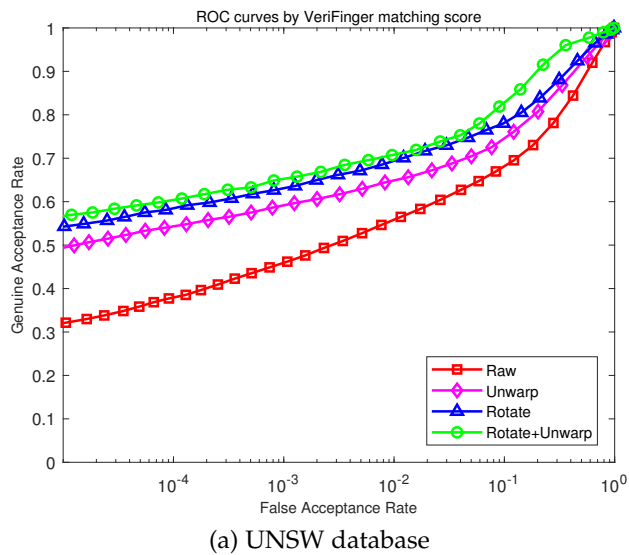}}
\centerline{(a) UNSW database}
\end{minipage}
\hfill
\begin{minipage}[b]{.45\linewidth}
\centering
  \centerline{\includegraphics[width=\linewidth]{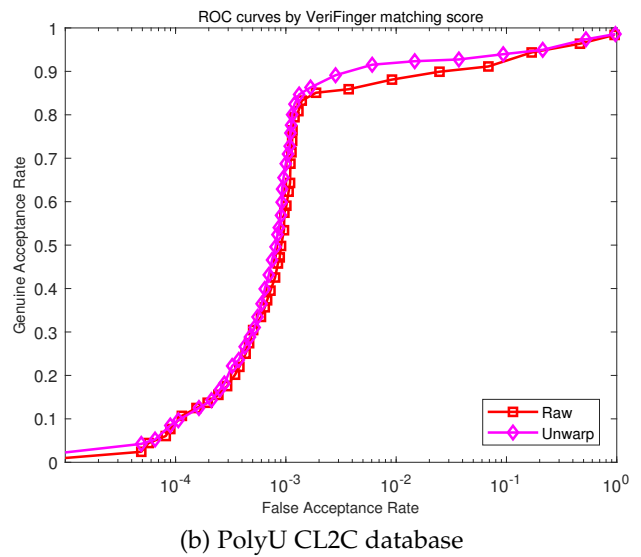}}
\centerline{(b) PolyU CL2C database}
\end{minipage}
\caption{ROC curves of contactless-contactless fingerprint matching on two databases.}
\label{fig:cmc2}
\end{figure*}
Fig. \ref{fig:cmc2} shows matching results of contactless-contactless fingerprints on UNSW and PolyU CL2C databases. The matching between contactless fingerprints does not use pose information to help matching. Namely, images of different poses are considered as independent samples. As we can see, unwarping increases matching performances.

For UNSW database, as input fingerprints and reference fingerprints both have three poses respectively, there are nine combinations of matching in total. The detailed matching results of these combinations are shown in Table \ref{table:eer2}. The proposed unwarping algorithm improves matching performances on all poses. Also, matching results on PolyU CL2C database are shown in Table \ref{table:eer11}. As we can see, unwarping improves the matching accuracy.

\begin{table}[!]
\caption{Equal error rates of contactless-contactless matching on UNSW database}
\centering
\begin{tabular}{ccccc}
\toprule
Pose& Raw&Unwarp&Rotate&\tabincell{c}{Rotate\\+Unwarp}\\
\midrule
Front-Front	&8.12	&\textbf{6.88}	&8.12	&\textbf{6.88}\\

Front-Right	&16.58	&13.38	&12.92	&\textbf{9.79}\\

Front-Left	&19.55	&16.75	&14.53	&\textbf{11.63}\\

Right-Front	&16.28	&13.12	&12.98	&\textbf{10.13}\\

Right-Right	&12.75	&10.97	&10.50	&\textbf{9.27}\\

Right-Left	&44.03	&38.27	&32.35	&\textbf{25.67}\\

Left-Front	&18.69	&16.73	&14.60	&\textbf{11.18}\\

Left-Right	&43.93	&37.13	&31.60	&\textbf{25.94}\\
 
Left-Left	&9.99	&8.80	&8.92	&\textbf{8.42}\\

All			&23.95	&20.18	&17.74	&\textbf{14.12}\\
\bottomrule
\end{tabular}
\label{table:eer2}
\end{table}

For contactless-contactless matching, we compare our method with the state-of-the-art method in \cite{tan2020towards}, which utilizes pose compensation to deal with perspective distortion. Table \ref{table:acc1} compares our matching result with \cite{tan2020towards} that both use VeriFinger \cite{verifinger} to compute the matching score. The results in Table \ref{table:acc1} are average matching results among all poses. As we can see, our method outperforms \cite{tan2020towards} in matching accuracy. As for PolyU CL2C database, there is currently no previous contactless-contactless matching algorithm on this database to be compared with.
\section{Limitations and Future Work}\label{sec:limitation}
The proposed 3D reconstruction and unwarping method can recover a 3D finger shape from a single contactless fingerprint image and obtain a rolled fingerprint without perspective distortion. Experimental results show that the proposed algorithm obtains good reconstruction accuracy and increases matching performances. But the current algorithm still has some limitations that deserve further exploration in the future.

\textbf{Reconstruction precision.} The current reconstruction algorithm recovers a smoothed finger shape without detailed ridge-valley structure. Although reconstructing 3D ridge-valley structure is a more powerful 3D reconstruction technology, a 3D shape with smoothed surface is sufficient for unwarping a contactless fingerprint to a rolled fingerprint. Even if 3D ridge-valley structure can be reconstructed, it needs to be converted back to 2D ridge-valley pattern for compatibility with existing 2D fingerprint systems. The ridge-valley structure is already conveyed by the 2D texture image. It is not clear yet if reconstructing 3D ridge-valley structure can bring benefit in real applications.

\textbf{Dataset precision.} The PolyU 3D and UNSW datasets used for training the gradient estimation network are of limited 3D accuracy, especially the UNSW dataset whose ground-truth depth map is reconstructed using images from three view angles by shape from silhouette method. But these databases are the most commonly used publicly available contactless fingerprint database. We have to utilize these databases to compare results with previous studies on these databases. Our method can benefit from more accurate 3D fingerprint. However, our method is already able to decrease perspective distortion and increase matching performance for contactless fingerprint even using 3D fingerprint data that are not so accurate. We plan to study how the accuracy of 3D training data affects the final matching performance.

\textbf{Dataset complexity.} The acquisition situations of contactless fingerprints in certain application scenarios may be more complex and flexible, such as collection by smart phone, raising higher requirements for the adaptability of the proposed algorithm. In this study, the proposed network was trained on two datasets where the camera parameters are not significantly different. Therefore, the algorithm with current parameters may have problems if directly applied to fingerprint photos that are taken from long distance. A simple solution is to collect dataset similar to the target application scenario and re-train the network. The two datasets in our study were collected in typical fingerprint recognition scenarios, and our method performed well on these normal situations. In the future, more training data under different imaging settings could be collected if we want to explore the generalization of our method to drastically different camera parameters.

\textbf{Scale ambiguity.} Generally speaking, it is very difficult to completely determine the scale for monocular 3D reconstruction method without changing the hardware or adding certain clues. Our solution is using the mean ridge period value to adjust the scale. However, the mean ridge periods for different people may be a little different. For example, the ridge period of a baby's fingerprint is smaller than an adult's fingerprint. Therefore, the reconstructed 3D fingerprint may have some scale ambiguity depending on the small differences between a certain finger's ridge period with average ridge period among all people. Although unifying ridge periods of all fingerprints does no harm to genuine matches, it may lead to increases in impostor matching scores. Much larger datasets are required to study this problem.

\textbf{Skin distortion.} The current algorithm solves perspective distortion for contactless fingerprints, but skin distortion also needs to be considered when matching with contact-based fingerprints. This paper deals with perspective distortion for contactless fingerprint, which is resulted from projecting a 3D finger into a 2D image. The proposed algorithm is effective for both contactless-contactless matching and contactless-contact matching as shown in our experiments. Because the skin distortion is very diverse depending on the direction and force of pressing finger, our unwarping method does not take skin distortion into consideration. To further improve the similarity between contactless fingerprint and contact-based fingerprint, distortion rectification for contact-based fingerprint can be studied \cite{Dabouei2019}\cite{si2015detection}.

\textbf{Feature extraction and matching.} The focus of this study is the geometric aspect of contactless fingerprints, while image quality also deserves attention. The feature extraction and matching methods used in matching experiment are still conventional methods developed for contacted-based fingerprints. Feature extraction and matching are certainly important for overall performance. We plan to study particular feature extraction and matching methods suitable for contactless fingerprints in the future.
\section{Conclusion}\label{sec:conclusion}
This paper proposes a new approach for contactless 3D fingerprint reconstruction and unwarping using deep learning. A contactless fingerprint is first preprocessed, then sent into the network to estimate surface gradients. The estimated gradients are used for 3D shape reconstruction, and finally the contactless fingerprint is unwarped. Experimental results on three databases show that the proposed method has low reconstruction error and high unwarping quality. Matching experiments prove the proposed unwarping method is able to reduce perspective distortion and is beneficial for contactless fingerprint matching.



%

%

\ifCLASSOPTIONcompsoc
  \section*{Acknowledgments}
\else
  \section*{Acknowledgment}
\fi

We are grateful to Dr. Feng Liu for providing us with the PolyU 3D database of contactless fingerprint images along with corresponding depth data. We would also like to thank those research groups who share their fingerprint data.

\ifCLASSOPTIONcaptionsoff
  \newpage
\fi


\bibliographystyle{IEEEtran}
\bibliography{ref}
\vfill


\end{document}